\def\cvprPaperID{4445}
\def\confName{CVPR}
\def\confYear{2023}

\def\paperTitle{PyPose: A Library for Robot Learning with Physics-based Optimization}

\def\authorBlock{Chen Wang$^{1,2,\textrm{\Letter}}$,
Dasong Gao$^{1,3}$,
Kuan Xu$^{4}$,
Junyi Geng$^1$,
Yaoyu Hu$^1$,
Yuheng Qiu$^1$,\\
Bowen Li$^1$,
Fan Yang$^5$,
Brady Moon$^1$,
Abhinav Pandey$^6$,
Aryan$^{1,7}$,
Jiahe Xu$^1$,
Tianhao Wu$^8$,\\
Haonan He$^1$,
Daning Huang$^6$,
Zhongqiang Ren$^1$,
Shibo Zhao$^1$,
Taimeng Fu$^9$,
Pranay Reddy$^{10}$,\\
Xiao Lin$^{11}$,
Wenshan Wang$^1$,
Jingnan Shi$^3$,
Rajat Talak$^3$,
Kun Cao$^{4}$,
Yi Du$^{2}$,
Han Wang$^4$,
Huai Yu$^{12}$,\\
Shanzhao Wang$^{13}$,
Siyu Chen$^{4}$,
Ananth Kashyap$^{14}$,
Rohan Bandaru$^{15}$,
Karthik Dantu$^2$,\\
Jiajun Wu$^{16}$,
Lihua Xie$^{4}$,
Luca Carlone$^{3}$,
Marco Hutter$^5$,
Sebastian Scherer$^1$\\
\url{https://pypose.org}\\
\thanks{$^{\textrm{\Letter}}$Corresponding Author. {\tt chenwang@dr.com}}
\thanks{$^1$Carnegie Mellon University, Pittsburgh, PA 15213, USA.}%
\thanks{$^2$State University of New York at Buffalo, NY 14260, USA.}
\thanks{$^3$Massachusetts Institute of Technology, Cambridge, MA 02139, USA.}
\thanks{$^4$Nanyang Technological University, Singapore 639798.}
\thanks{$^5$ETH Z\"urich, 8092 Z\"urich, Switzerland.}
\thanks{$^6$Pennsylvania State University, University Park, PA 16801, USA.}
\thanks{$^7$Delhi Technological University, Delhi, India.}
\thanks{$^8$University of Virginia, Charlottesville, VA 22904, USA.}
\thanks{$^9$The Chinese University of Hong Kong, Shenzhen, China.}
\thanks{$^{10}$University of Massachusetts Amherst, MA 01003, USA.}
\thanks{$^{11}$Georgia Institute of Technology, Atlanta, GA 30332, USA}
\thanks{$^{12}$Wuhan University, Hubei 430072, China.}
\thanks{$^{13}$University of Michigan, Ann Arbor, MI 48109, USA.}
\thanks{$^{14}$Fox Chapel Area High School, Pittsburgh, PA 15238, USA.}
\thanks{$^{15}$Lexington High School, Lexington, MA 02421, USA.}
\thanks{$^{16}$Stanford University, Stanford, CA 94305, USA.}
}

\makeatletter
\def\thanks#1{\protected@xdef\@thanks{\@thanks
        \protect\footnotetext{#1}}}
\makeatother

\newif\ifreview 
\newif\ifarxiv \newcommand{\arxiv}{\arxivtrue}
\newif\ifcamera 
\newif\ifrebuttal 

\arxiv 
\pdfoutput=1
\documentclass[10pt,twocolumn,letterpaper]{article}
\ifreview \usepackage[review]{cvpr} \fi
\ifarxiv \usepackage[pagenumbers]{cvpr} \fi
\ifrebuttal \usepackage[rebuttal]{cvpr} \fi
\ifcamera \usepackage{cvpr} \fi

\usepackage{graphicx}
\usepackage{amsmath}
\usepackage{amssymb}
\usepackage{booktabs}

\usepackage{times}
\usepackage{microtype}
\usepackage{epsfig}
\usepackage[table,xcdraw]{xcolor}
\usepackage{caption}
\usepackage{float}
\usepackage{placeins}
\usepackage{color, colortbl}
\usepackage{stfloats}
\usepackage{tabularx}
\usepackage{xstring}
\usepackage{multirow}
\usepackage{xspace}
\usepackage{url}
\usepackage{subcaption}
\usepackage{xcolor}
\usepackage[hang,flushmargin]{footmisc}

\ifcamera \usepackage[accsupp]{axessibility} \fi

% ...

\ifarxiv  \fi

\newcommand{\R}[1]{{%
    \textbf{%
        \ifstrequal{#1}{1}{\textcolor{red}{R#1}}{%
        \ifstrequal{#1}{2}{\textcolor{blue}{R#1}}{%
        \ifstrequal{#1}{3}{\textcolor{magenta}{R#1}}{%
        \ifstrequal{#1}{4}{\textcolor{teal}{R#1}}{%
                           \textcolor{cyan}{R#1}%
        }}}}%
    }%
}}

\usepackage{xr-hyper}

\makeatletter
\newcommand*{\addFileDependency}[1]{
  \typeout{(#1)}
  \@addtofilelist{#1}
  \IfFileExists{#1}{}{\typeout{No file #1.}}
}

\makeatother

\usepackage[pagebackref,breaklinks,colorlinks]{hyperref}
\usepackage[capitalize]{cleveref}
\crefname{section}{Sec.}{Secs.}
\crefname{table}{Table}{Tables}
\crefname{figure}{Fig.}{Figs.}

\hypersetup{
pdftitle={PyPose: A Library for Robot Learning with Physics-based Optimization},
pdfsubject={Robotics},
pdfauthor={Chen Wang, Dasong Gao, Kuan Xu, Junyi Geng, Yaoyu Hu, Yuheng Qiu, Bowen Li, Fan Yang, Brady Moon, Abhinav Pandey, Aryan, Jiahe Xu, Tianhao Wu, Haonan He, Daning Huang, Zhongqiang Ren, Shibo Zhao, Taimeng Fu, Pranay Reddy, Xiao Lin, Wenshan Wang, Jingnan Shi, Rajat Talak, Kun Cao, Yi Du, Han Wang, Huai Yu, Shanzhao Wang, Siyu Chen, Ananth Kashyap, Rohan Bandaru, Karthik Dantu, Jiajun Wu, Lihua Xie, Luca Carlone, Marco Hutter, Sebastian Scherer},
pdfkeywords={Robotics, Perception, Physics-based Optimization, Open Source}
}

\usepackage[accsupp]{axessibility}  
\usepackage[T1]{fontenc} 

\usepackage{amsthm}
\usepackage{amsmath}
\usepackage{amssymb}
\usepackage{commath}
\usepackage{bm} 
\usepackage{multirow}
\usepackage{enumerate}
\usepackage{enumitem}
\usepackage{wrapfig}
\usepackage{pifont}%
\usepackage{footnote}
\usepackage{balance}
\usepackage{threeparttable}
\usepackage[amssymb]{SIunits}
\usepackage{epsfig}
\usepackage{graphicx}
\usepackage{subcaption}
\usepackage[super]{nth}
\usepackage{pythonhighlight}
\usepackage[misc,geometry]{ifsym}

\usepackage[font={small}]{caption}

\usepackage{amssymb}
\usepackage{pifont}
\usepackage{url}            
\usepackage{booktabs}       
\usepackage{amsfonts}       
\usepackage{nicefrac}       
\usepackage{microtype}      
\usepackage{lipsum}

\usepackage{physics}
\usepackage{tabularx}
\usepackage[normalem]{ulem}
\usepackage[]{graphicx}

\newsavebox{\measurebox}

\makeatletter
\let\NAT@parse\undefined
\makeatother

\useunder{\uline}{\ul}{}

\newcommand{\fref}[1]{Figure~\ref{#1}}
\newcommand{\sref}[1]{Section~\ref{#1}}
\newcommand{\tref}[1]{Table~\ref{#1}}

\usepackage{array}
\newcommand{\PreserveBackslash}[1]{\let\temp=\\#1\let\\=\temp}
\newcolumntype{C}[1]{>{\PreserveBackslash\centering}p{#1}}
\newcolumntype{R}[1]{>{\PreserveBackslash\raggedleft}p{#1}}
\newcolumntype{L}[1]{>{\PreserveBackslash\raggedright}p{#1}}

\begin{document}
\title{\paperTitle}
\author{\authorBlock}
\maketitle

\begin{abstract}
\vspace{-3mm}
Deep learning has had remarkable success in robotic perception, but its data-centric nature suffers when it comes to generalizing to ever-changing environments.
By contrast, physics-based optimization generalizes better, but it does not perform as well in complicated tasks due to the lack of high-level semantic information and reliance on manual parametric tuning.
To take advantage of these two complementary worlds, we present PyPose: a robotics-oriented, PyTorch-based library that combines deep perceptual models with physics-based optimization.
PyPose's architecture is tidy and well-organized, it has an imperative style interface and is efficient and user-friendly, making it easy to integrate into real-world robotic applications.
Besides, it supports parallel computing of any order gradients of Lie groups and Lie algebras and \nth{2}-order optimizers, such as trust region methods.
Experiments show that PyPose achieves more than $10\times$ speedup in computation compared to the state-of-the-art libraries.
To boost future research, we provide concrete examples for several fields of robot learning, including SLAM, planning, control, and inertial navigation. \ifreview The source code will be made public.\fi
\end{abstract}
\vspace{-5mm}
\section{Introduction} \label{sec:intro}
\vspace{-2mm}
Deep learning has made great inroads in visual perception tasks such as classification \cite{wang2019kervolutional}, segmentation \cite{minaee2021image}, and detection \cite{li2021airdet}.
However, it is still often unsatisfactory in some robotic applications due to the lack of training data \cite{wang2021unsupervised}, ever-changing environments \cite{zhao2021super}, and limited computational resources \cite{wang2017non}.
On the other hand, physics-based optimization has shown great generalization ability and high accuracy in many vision and robotic tasks, such as control \cite{fang2020cooperative}, planning \cite{yang2021far}, and simultaneous localization and mapping (SLAM) \cite{zhao2021super}. 
Nevertheless, it relies on problem-specific parameter tuning and suffers from the lack of semantic information.
Since both methods have shown their own merits, more and more efforts have been made to take advantage of the two complementary worlds \cite{zhao2020tp}.

Currently, learning-based methods and physics-based optimization are typically used separately in different modules of a robotic system \cite{ebadi2022present}.
For example, in semantic SLAM, learning-based methods showed promising results in scenarios where high-level semantic information is needed or as a replacement for hand-crafted features and descriptors, e.g., feature matching in the front-end \cite{sarlin2020superglue}, while physics-based optimization plays a vital role in cases where a well-defined physical model can be established, e.g., pose graph optimization in the back-end \cite{campos2021orb}. 
Researchers usually first execute the front end and then pass the results to the back end for optimization.
Despite the tremendous progress in the past decades, such a two-stage, decoupled paradigm may only achieve sub-optimal solutions, which in turn limits system performance and generalization ability. Hence, developing integrated methods with end-to-end differentiation through optimization is an emerging trend \cite{teed2021droid, teed2021tangent, teed2022deep}.

A variety of applications in perception, motion planning, and automatic control have been explored  for end-to-end learning \cite{teed2021droid,hafner2019learning,lenz2015deepmpc}.  However, most of these applications rely on problem-specific implementations that are often coded from scratch, which makes it difficult for researchers to build upon prior work and explore new ideas. This hinders the development cycle due to the lack of a unified and systematic development framework. For example, people usually leverage PyTorch-based \cite{paszke2019pytorch} models for developing learning-based perceptual models, but then have to use  C(++)-based optimization libraries, such as GTSAM \cite{dellaert2012factor}, OMPL \cite{sucan2012open}, and CT \cite{giftthaler2018control}, for physics-based optimization.
The mixed usage of Python and C++ libraries increases the system complexity and slows down the development cycle as it is time-consuming for cross-language debugging and inefficient to transfer data among different processes, e.g., ROS nodes \cite{quigley2009ros}.
Therefore, there is an urgent need for a systematic development tool in a single language, accelerating end-to-end learning for physics-based optimization.

Some researchers have spent effort towards this objective. For example, LieTorch exploits the group structure of 3D transformations and performs back-propagation in the tangent spaces of manifolds~\cite{teed2021tangent}. However, only \nth{1}-order differentiable operations are currently implemented, which limits its practical use, since higher order derivatives provide additional local information about the data distribution and enable new applications \cite{meng2021estimating}.
CvxpyLayer \cite{cvxpylayers2019} takes convex optimization as a differentiable neural network layer, while it doesn't support operation for Lie groups and \nth{2}-order optimizers.
Similarly, Theseus \cite{pineda2022theseus} takes non-linear optimization as network layers; however, it adopts rotation matrices for transformation representation, which is memory inefficient for practical robotic applications.

To address the above limitations, we present PyPose, an open-source library based on PyTorch to connect learning-based perceptual models with classical algorithms that can be formulated as physics-based optimization, e.g., geometry problem, factor-graph optimization, and optimal control.
In summary, our main contributions are:
\begin{itemize}[noitemsep,topsep=0pt]
    \item We present a new python-based open-source library, PyPose, to further enable end-to-end learning with physics-based optimization and accelerate the next generation of developments in robotics. PyPose is designed to be easily interpretable, user-friendly, and efficient with a tidy and well-organized architecture.
    It provides an imperative programming style for the convenience of real-world robotic applications.
    PyPose supports any order gradient computation of Lie groups and Lie algebras, and \nth{2}-order optimizers such as Levenberg-Marquardt with trust region steps. As demonstrated in \fref{fig:jacobian}, our experiments show that PyPose achieves more than $10\times$ faster compared to state-of-the-art libraries.
    \item We provide sample uses of PyPose. Users can easily build upon existing functionalities for various robotic applications. To the best of our knowledge, PyPose is one of the first Python libraries to comprehensively cover several sub-fields of robotics, such as perception, SLAM, and control, where optimization is involved.
\end{itemize}

\vspace{-2mm}
\section{Related Work}
\label{sec:related}
\vspace{-2mm}
Existing open source libraries related to PyPose can mainly be divided into two groups: (1) linear algebra and machine learning frameworks and (2) optimization libraries.

\subsection{Linear Algebra \& Machine Learning}
\vspace{-2mm}
Linear algebra libraries are essential to machine learning and robotics research.
To name a few, NumPy \cite{oliphant2006guide}, a linear algebra library for Python, offers comprehensive operations on vectors and matrices while enjoying higher running speed due to its underlying well-optimized C code.
Eigen \cite{guennebaud2010eigen}, a high performance C++ linear algebra library, has been used in many projects such as TensorFlow \cite{abadi2016tensorflow}, Ceres \cite{AgarwalCeresSolver2022}, GTSAM \cite{dellaert2012factor}, and g$^2$o \cite{grisetti2011g2o}.
ArrayFire \cite{malcolm2012arrayfire}, a GPU acceleration library for C, C++, Fortran, and Python, contains simple APIs and provides thousands of GPU-tuned functions.

While relying heavily on high-performance numerical linear algebra techniques, machine learning libraries focus more on operations on tensors (i.e., high-dimensional matrices) and automatic differentiation.
Early machine learning frameworks, such as Torch \cite{collobert2002torch}, OpenNN \cite{open2016open}, and MATLAB \cite{MATLAB2010}, provide primitive tools for researchers to develop neural networks.
However, they only support CPU computation and lack concise APIs, which plague engineers using them in applications.
A few years later, deep learning frameworks such as Chainer \cite{tokui2015chainer}, Theano \cite{al2016theano}, and Caffe \cite{jia2014caffe} arose to handle the increasing size and complexity of neural networks while supporting multi-GPU training with convenient APIs for users to build and train their neural networks.
Furthermore, the recent frameworks, such as TensorFlow \cite{abadi2016tensorflow}, PyTorch \cite{paszke2019pytorch}, and MXNet \cite{chen2015mxnet}, provide a comprehensive and flexible ecosystem (e.g., APIs for multiple programming languages, distributed data parallel training, and facilitating tools for benchmark and deployment). 
Gvnn \cite{handa2016gvnn} introduced differentiable transformation layers into Torch-based framework, leading to end-to-end geometric learning.
JAX \cite{jax2018github} can automatically differentiate native Python and NumPy functions and is an extensible system for composable function transformations.
In many ways, the existence of these frameworks facilitated and promoted the growth of deep learning.
Recently, more efforts have been taken to combine standard optimization tools with deep learning. 
Recent work like Theseus \cite{pineda2022theseus} and CvxpyLayer \cite{cvxpylayers2019} showed how to embed differentiable optimization within deep neural networks.

By leveraging PyTorch, our proposed library, PyPose, enjoys the same benefits as the existing state-of-the-art deep learning frameworks.
Additionally, PyPose provides new features, such as \nth{2}-order optimizers and  computing any order gradients of Lie groups and Lie algebras, which are essential to geometrical optimization problems and robotics.

\subsection{Open Source Optimization Libraries}
\vspace{-2mm}
Numerous optimization solvers and frameworks have been developed and leveraged in robotics.
To mention a few, Ceres \cite{AgarwalCeresSolver2022} is an open-source C++ library for large-scale nonlinear least squares optimization problems and has been widely used in SLAM. 
Pyomo \cite{hart2017pyomo} and JuMP \cite{dunning2017jump} are optimization frameworks that have been widely used due to their flexibility in supporting a diverse set of tools for constructing, solving, and analyzing optimization models.
CasADi \cite{andersson2019casadi} has been used to solve many real-world control problems in robotics due to its fast and effective implementations of different numerical methods for optimal control.

Pose- and factor-graph optimization also play an important role in robotics.
For example, g$^2$o \cite{grisetti2011g2o} and GTSAM \cite{dellaert2012factor} are open-source C++ frameworks for graph-based nonlinear optimization, which provide concise APIs for constructing new problems and have been leveraged to solve several optimization problems in SLAM and bundle adjustment.

\begin{table}[t]
    \small
    \caption{A list of supported class alias of \texttt{LieTensor}.}
    \vspace{-5pt}
    \label{tab:lietensor}
    \centering
    \resizebox{1\linewidth}{!}{
    \begin{tabular}{C{0.48\linewidth}C{0.18\linewidth}C{0.18\linewidth}}
        \toprule
        Transformation &  Lie Group & Lie Algebra \\\midrule
        Rotation & \texttt{SO3()}	& \texttt{so3()}  \\
        Rotation \& Translation & \texttt{SE3()} & \texttt{se3()} \\
        Rotation \& Translation \& Scale &  \texttt{Sim3()} & \texttt{sim3()} \\
        Rotation \& Scale & \texttt{RxSO3()} & \texttt{rxso3()} \\
        \bottomrule
    \end{tabular}
    }
    \vspace{-10pt}
\end{table}

Optimization libraries have also been widely used in robotic control problems.
To name a few, IPOPT \cite{wachter2006implementation} is an open-source C++ solver for nonlinear programming problems based on interior-point methods and is widely used in robotics and control.
Similarly, OpenOCL \cite{koenemann2017openocl} supports a large class of optimization problems such as continuous time, discrete time, constrained, unconstrained, multi-phase, and trajectory optimization problems for real-time model-predictive control.
Another library for large-scale optimal control and estimation problems is CT \cite{giftthaler2018control}, which provides standard interfaces for different optimal control solvers and can be extended to a broad class of dynamical systems in robotic applications.
In addition, Drake \cite{drake} also has solvers for common control problems and that can be directly integrated with its simulation tool boxes. Its system completeness made it favorable to many researchers.

\vspace{-2mm}
\section{Method}
\label{sec:method}
\vspace{-2mm}
To reduce the learning curve of new users, we leverage---to the greatest possible extent---existing structures in PyTorch.
Our design principles are to keep the implementation logical, modular, and transparent so that users can focus on their own applications and 
invest effort in the mathematical aspects of their problems rather than the implementation details.
Similar to PyTorch, we believe that trading a bit of efficiency for interpretability and ease of development is acceptable. However, we
are still careful in maintaining a good balance between interpretability and efficiency by leveraging advanced PyTorch functionalities.

Following these principles, we mainly provide four concepts to enable end-to-end learning with physics-based optimization, i.e., \texttt{LieTensor}, \texttt{Module}, \texttt{Function}, and \texttt{Optimizer}.
We briefly present their motivation, mathematical underpinning, and the interfaces in PyPose.

\subsection{\texttt{LieTensor}}
\vspace{-2mm}
In robotics, 3D transformations are crucial for many applications, such as SLAM, control, and motion planning. However, most machine learning libraries, including PyTorch, assume that the computation graph operates in Euclidean space, while a 3D transformation lies on a smooth manifold. 
Simply neglecting the manifold structure will lead to inconsistent gradient computation and numerical issues.
Therefore, we need a specific data structure to represent 3D transformations in learning models.

\begin{figure*}
	\centering
	\subfloat{\includegraphics[width=0.166\linewidth]{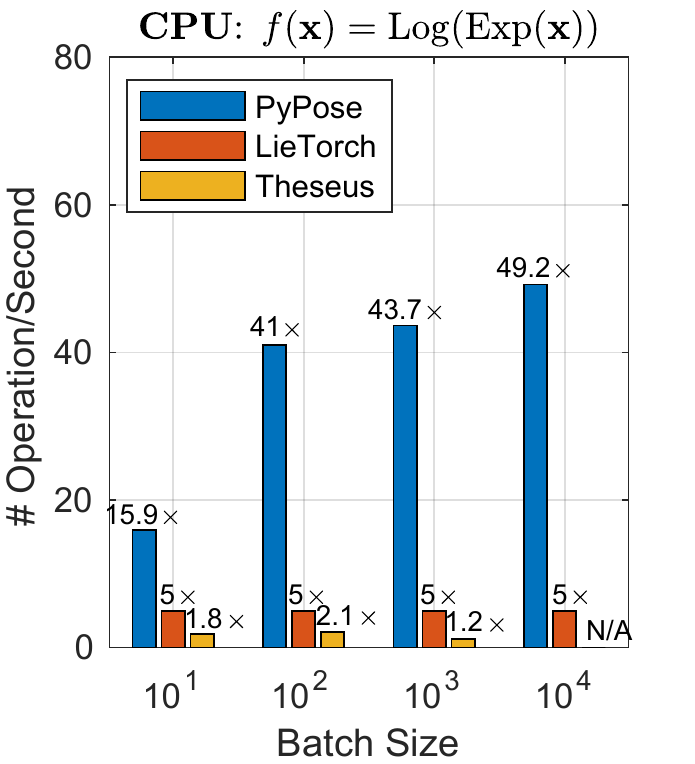}}
	\hfill
	\subfloat{\includegraphics[width=0.166\linewidth]{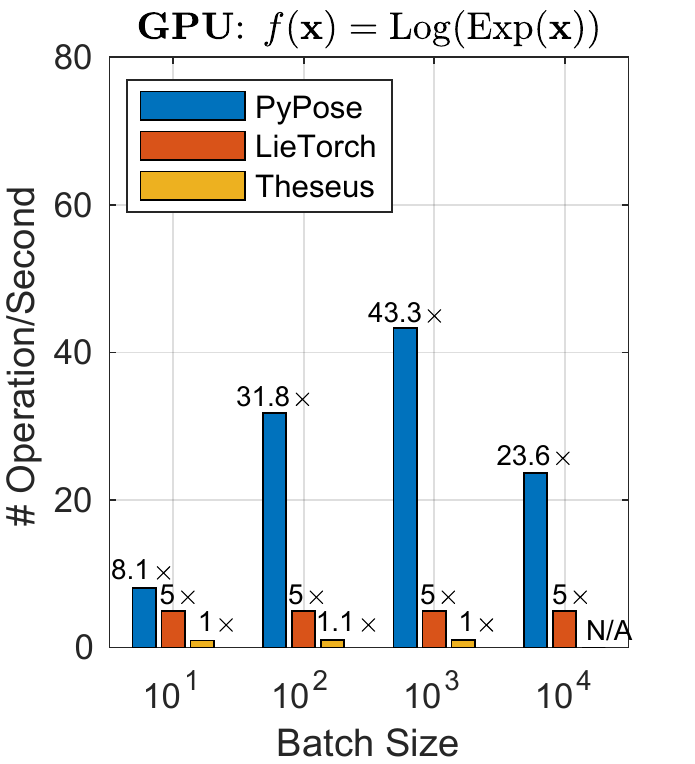}}
	\hfill
	\subfloat{\includegraphics[width=0.166\linewidth]{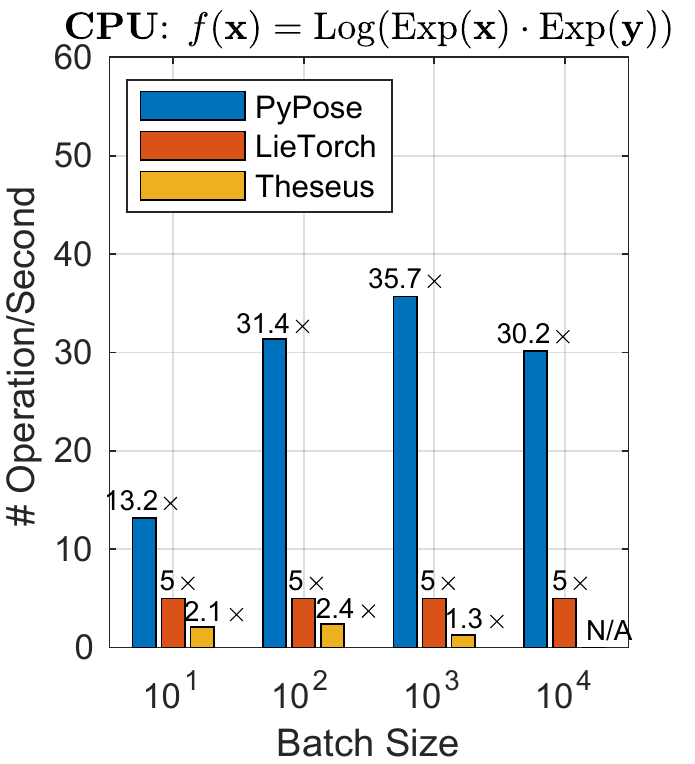}}
	\hfill
	\subfloat{\includegraphics[width=0.166\linewidth]{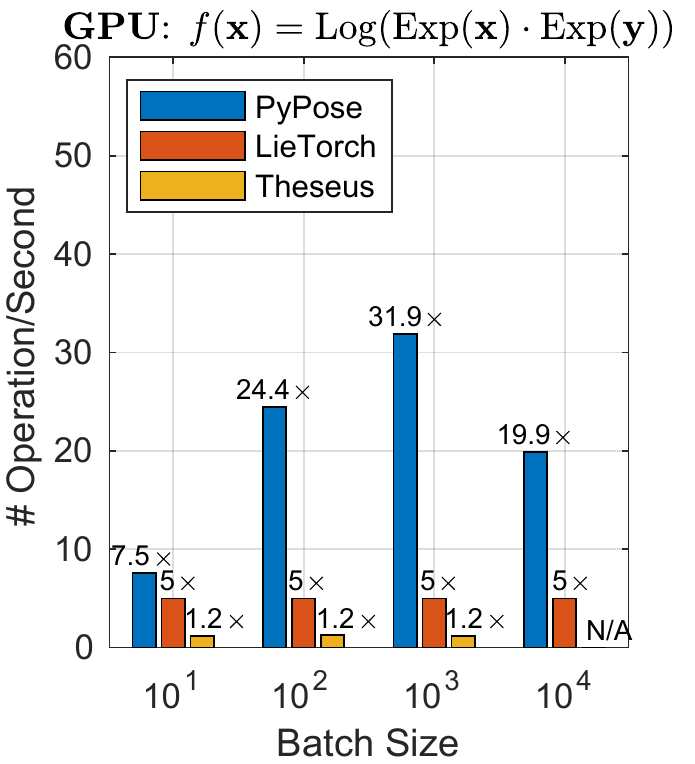}}
	\hfill
	\subfloat{\includegraphics[width=0.166\linewidth]{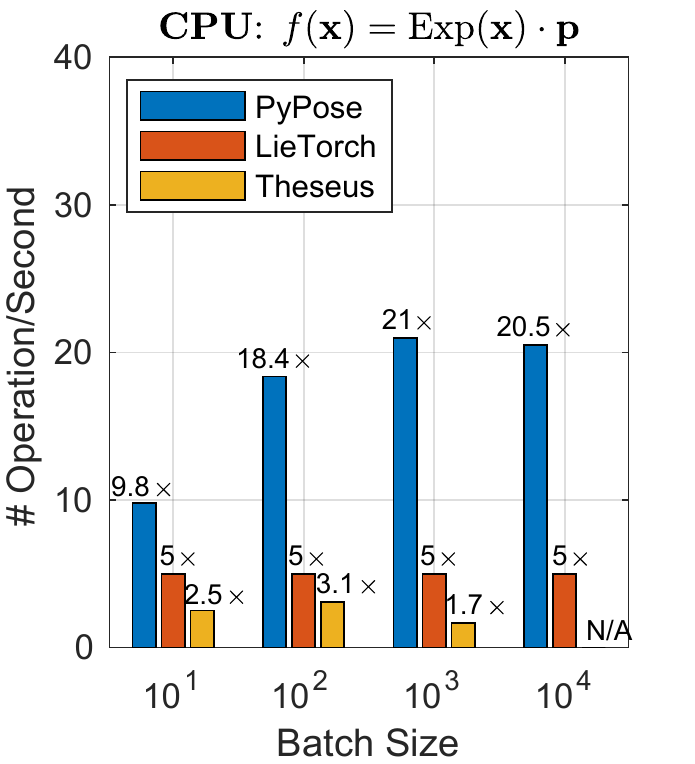}}
	\hfill
	\subfloat{\includegraphics[width=0.166\linewidth]{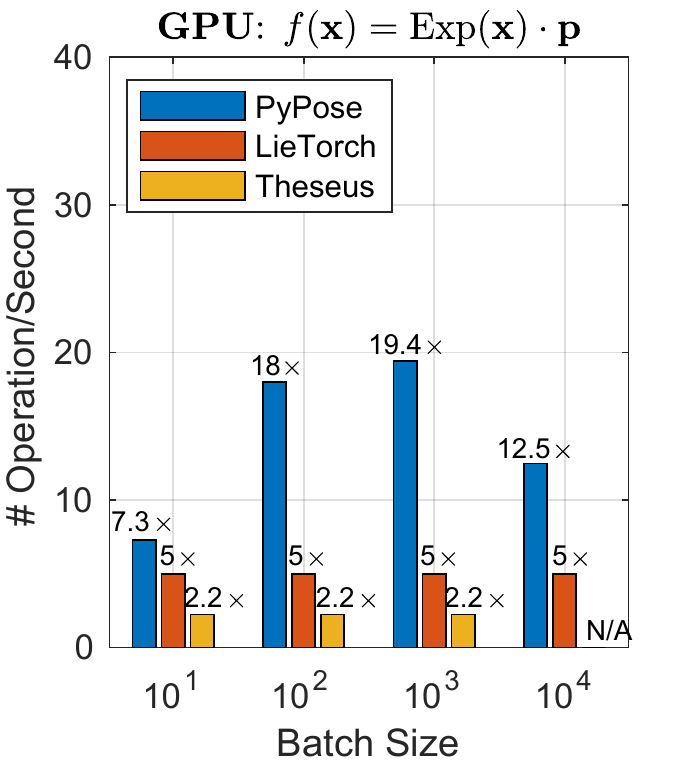}}
 
    \vspace{-5pt}
    \caption{Efficiency comparison of dense Jacobian for Lie group operations on CPU/GPU, where ``N/A'' means performance is unavailable with that batch size. The performance of Theseus \cite{pineda2022theseus} is computed with its \texttt{AutoDiffCostFunction} method. We take LieTorch performance as $5\times$ to be the baseline since Theseus runs out of memory at the batch size of $10^4$.
	}
	\label{fig:jacobian}
	\vspace{-10pt}
\end{figure*}

To address the above challenge while keeping PyTorch's excellent features, we resort to Lie theory and define \texttt{LieTensor} as a subclass of PyTorch's \texttt{Tensor} to represent 3D transformations.
One of the challenges of implementing a differentiable \texttt{LieTensor} is that we often need to calculate numerically problematic terms such as $\frac{\sin x}{x}$ for the conversion between a Lie group and Lie algebra \cite{teed2021tangent}. 
To solve this problem, we take the Taylor expansion to avoid calculating the division by zero.
To illustrate this, consider the exponential map from $\mathbb{R}^3$ to $\mathbb{S}^3$ as an example, where $\mathbb{S}^3$ is the set of  
unit-quaternions.
For numerical stability, the exponential map from $\mathbb{R}^3$ to $\mathbb{S}^3$ is formulated as follows:
\begin{equation*}
    \texttt{Exp}(\bm{x}) = \left\{
    \begin{aligned}
    &\left[\bm{x}^T\gamma_e,~ \cos(\frac{\|\bm{x}\|}{2})\right]^T & \|\bm{x}\| > \text{eps}\\
    &\left[\bm{x}^T\gamma_o,~
    1 - \frac{\|\bm{x}\|^2}{8} + \frac{\|\bm{x}\|^4}{384} \right]^T & \text{otherwise},\\
    \end{aligned}
    \right.
\end{equation*}
where $\gamma_e = \frac{1}{\|\bm{x}\|}\sin(\frac{\|\bm{x}\|}{2})$, $\gamma_o = \frac{1}{2} - \frac{1}{48} \|\bm{x}\|^2 + \frac{1}{3840} \|\bm{x}\|^4$, and $\text{eps}$ is the smallest machine number where $1 + \text{eps} \ne 1$.

\texttt{LieTensor} is different from the existing libraries in several aspects:
(\textbf{1}) PyPose supports auto-diff for any order gradient and is compatible with most popular devices, such as CPU, GPU, TPU, and Apple silicon GPU, while other libraries like LieTorch \cite{teed2021tangent} implement customized CUDA kernels and only support \nth{1}-order gradient.
(\textbf{2}) \texttt{LieTensor} supports parallel computing of gradient with the \texttt{vmap} operator, which allows it to compute Jacobian matrices much faster.
(\textbf{3})~Theseus \cite{pineda2022theseus} adopts rotation matrices, which require more memory, while PyPose uses quaternions, which only require storing four scalars, have no gimbal lock issues, and have better numerical properties.
(\textbf{4})~Libraries such as LieTorch, JaxLie \cite{jaxlie}, and Theseus only support Lie groups, while PyPose supports both Lie groups and Lie algebras. As a result, one can directly call the \texttt{Exp} and \texttt{Log} maps from a \texttt{LieTensor} instance, which is more flexible and user-friendly. Moreover, the gradient with respect to both types can be automatically calculated and back-propagated.

For other types of supported 3D transformations, the reader may refer to \tref{tab:lietensor} for more details.
The readers may find a list of supported \texttt{LieTensor} operations in the documentation.\footnote{\href{https://pypose.org/docs/main/basics}{https://pypose.org/docs/main/basics}}
A sample code for how to use a \texttt{LieTensor} is given in the supplementary and PyPose tutorials.\footnote{\href{https://pypose.org/tutorials/beginner/started}{https://pypose.org/tutorials/beginner/started}}

\vspace{-2mm}
\subsection{\texttt{Module} and \texttt{Function}}
\vspace{-2mm}
PyPose leverages the existing concept of \texttt{Function} and \texttt{Module} in PyTorch to implement differentiable robotics-related functionalities.
Concretely, a \texttt{Function} performs a specific computation given inputs and returns outputs; a \texttt{Module} has the same functionality as a \texttt{Function}, but it also stores data, such as \texttt{Parameter}, which is of type \texttt{Tensor} or \texttt{LieTensor}, that can be updated by an optimizer as discussed in the next section.

PyPose provides many useful modules, such as the system transition function, model predictive control (MPC), Kalman filter, and IMU preintegration. 

A list of supported \texttt{Module} can be found in the documentation.\footnote{\href{https://pypose.org/docs/main/modules}{https://pypose.org/docs/main/modules}}
Users can easily integrate them into their own systems and perform a specific task, e.g., a \texttt{System} module can be used by both \texttt{EKF} and \texttt{MPC}.
A few practical examples in SLAM, planning, control, and inertial navigation are presented in \sref{sec:examples}.

\vspace{-2mm}
\subsection{\texttt{Optimizer}}
\vspace{-2mm}
To enable end-to-end learning with physics-based optimization, we need to integrate general optimizers beyond the basic gradient-based methods such as SGD  \cite{robbins1951stochastic} and Adam \cite{kingma2014adam}, since many practical problems in robotics require more advanced optimization such as constrained or \nth{2}-order optimization \cite{barfoot2017state}.
In this section, we take a robust non-linear least square problem as an example and present the intuition behind the optimization-oriented interfaces of PyPose, including \texttt{solver}, \texttt{kernel}, \texttt{corrector}, and \texttt{strategy} for using the \nth{2}-order Levenberg-Marquardt (\texttt{LM}) optimizer.
Consider a weighted least square problem:
\begin{equation}\label{eq:least-square}
    \min_{\bm{\theta}} \sum_i 
            \left(\bm{f}(\bm{\theta},\bm{x}_i)-\bm{y}_i\right)^T \mathbf{W}_i
            \left(\bm{f}(\bm{\theta},\bm{x}_i)-\bm{y}_i\right),
\end{equation}
where $\bm{f}(\cdot)$ is a regression model (\texttt{Module}), $\bm{\theta}\in\mathbb{R}^n$ is the parameters to be optimized, $\bm{x}_i\in\mathbb{R}^m$ is an input, $\bm{y}_i\in\mathbb{R}^d$ is the target, $\mathbf{W}_i\in\mathbb{R}^{d\times d}$ is a square information matrix.

\begin{figure*}
	\centering
	\subfloat[CPU runtime.]{\includegraphics[width=0.2\linewidth]{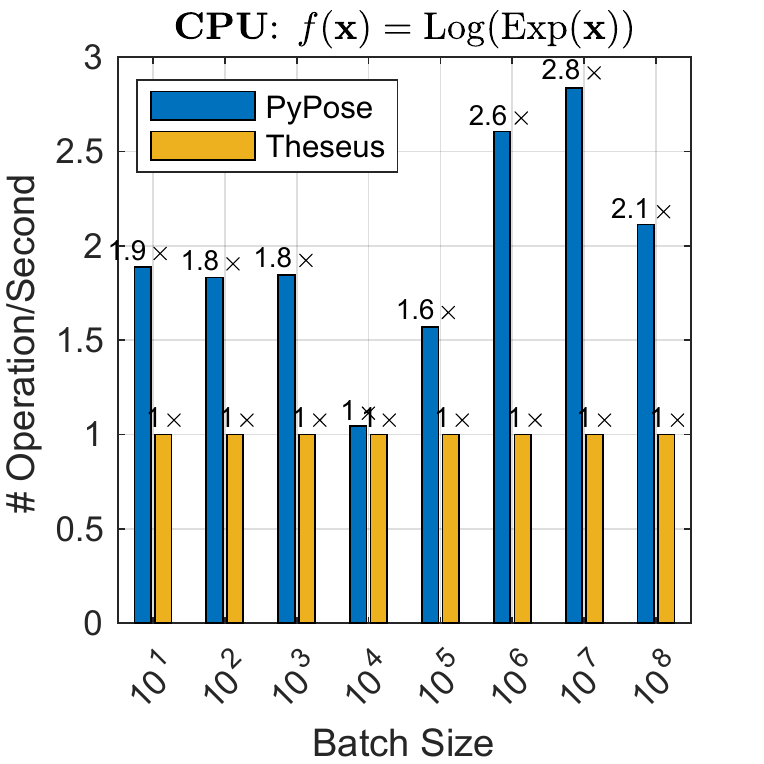}}
	\hfill
	\subfloat[GPU runtime.]{\includegraphics[width=0.2\linewidth]{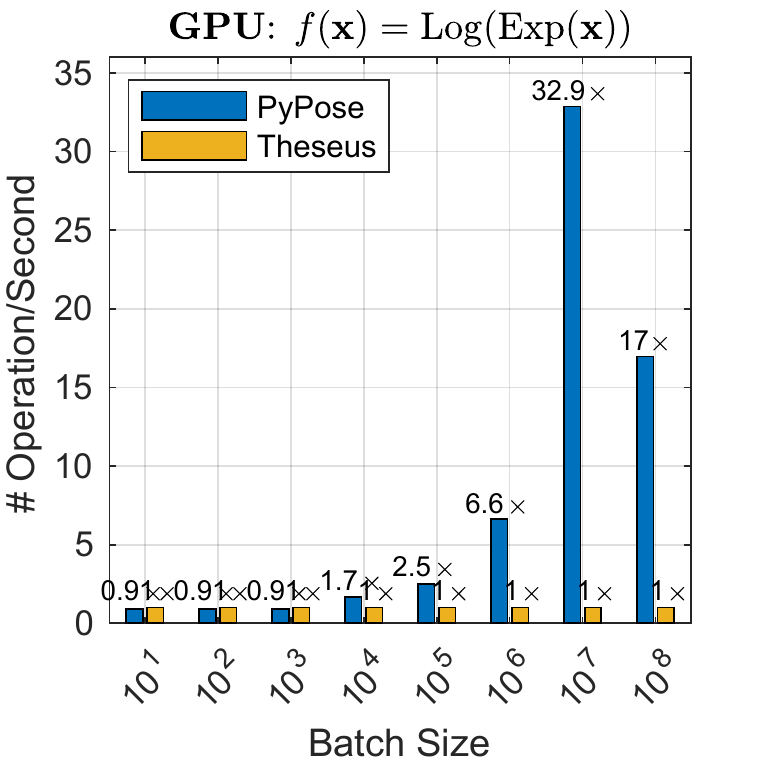}}
	\hfill
	\subfloat[CPU Jacobian runtime.]{\includegraphics[width=0.2\linewidth]{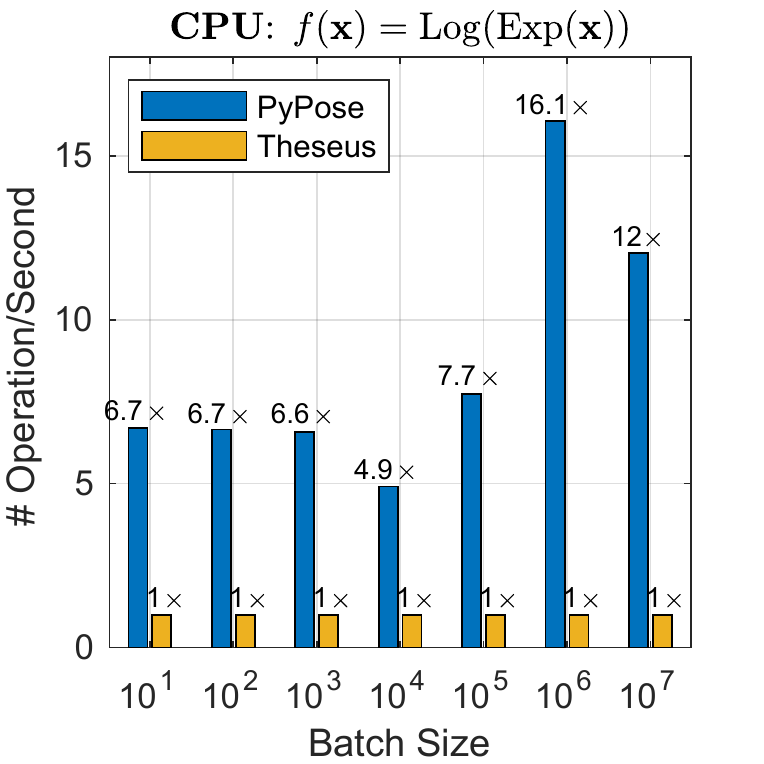}}
	\hfill
	\subfloat[GPU Jacobian runtime.]{\includegraphics[width=0.2\linewidth]{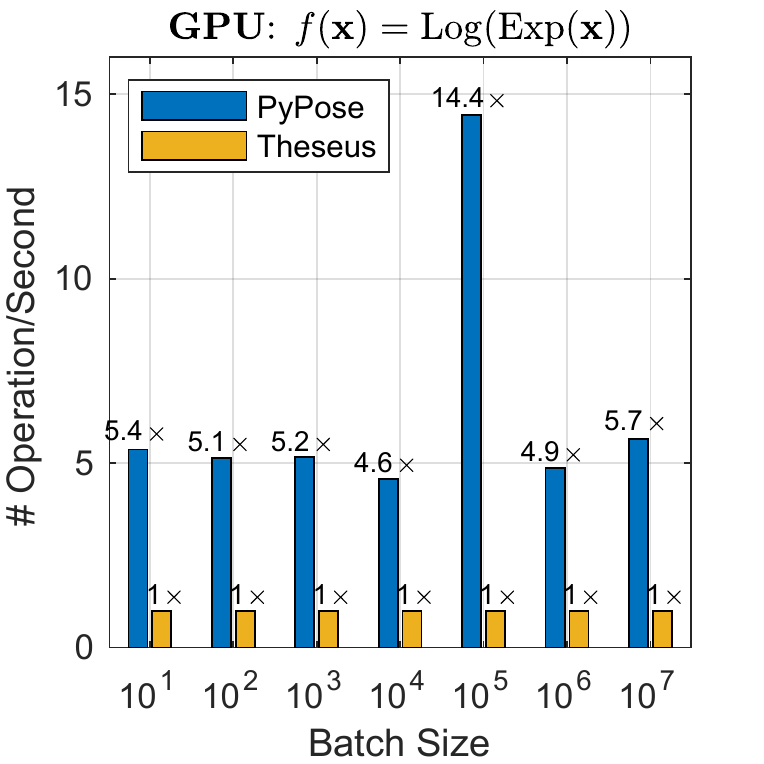}}
	\hfill
	\subfloat[Memory usage.]{\includegraphics[width=0.2\linewidth]{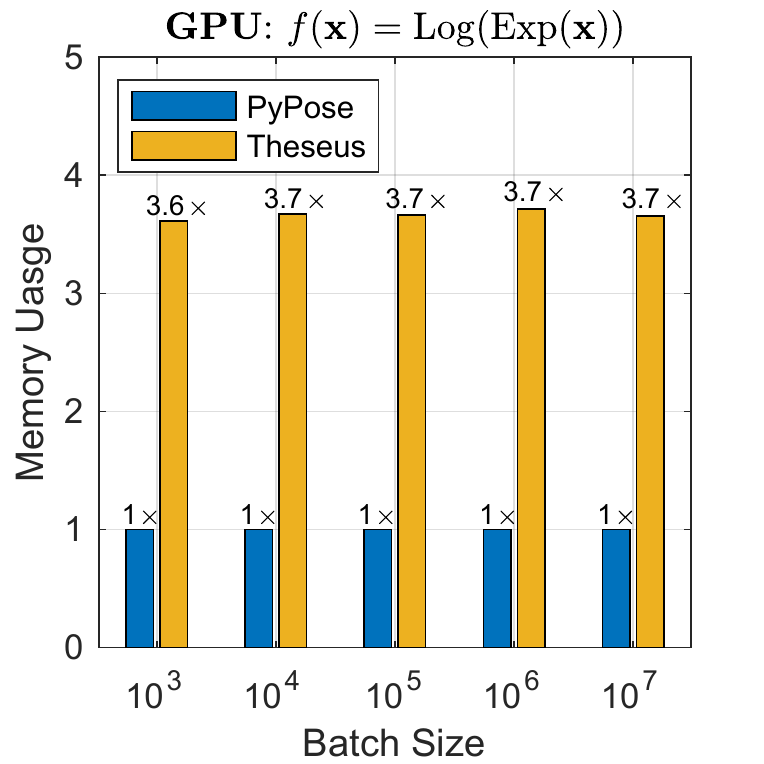}}
	\vspace{5pt}
	\subfloat[CPU runtime.]{\includegraphics[width=0.2\linewidth]{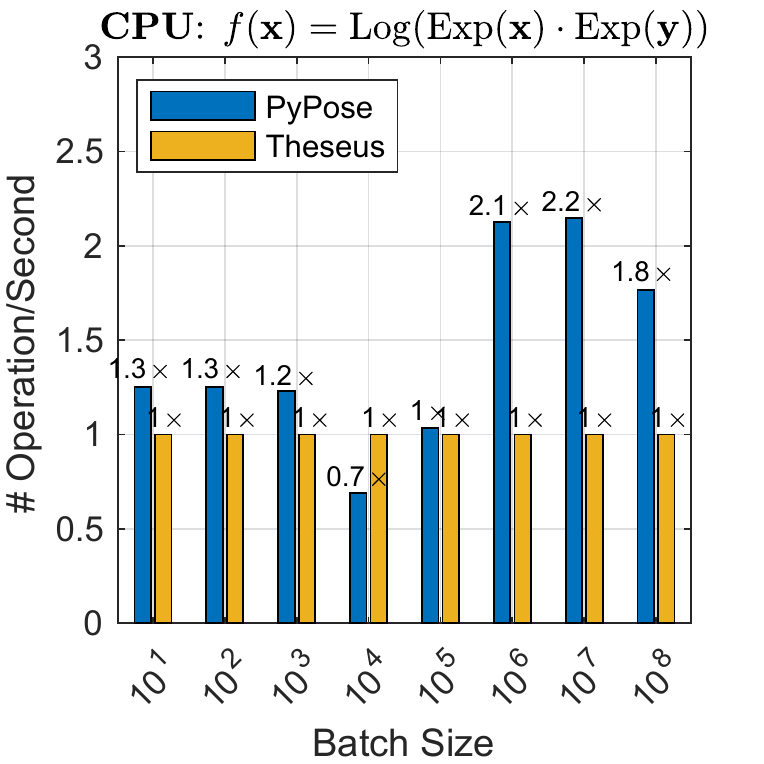}}
	\hfill
	\subfloat[GPU runtime.]{\includegraphics[width=0.2\linewidth]{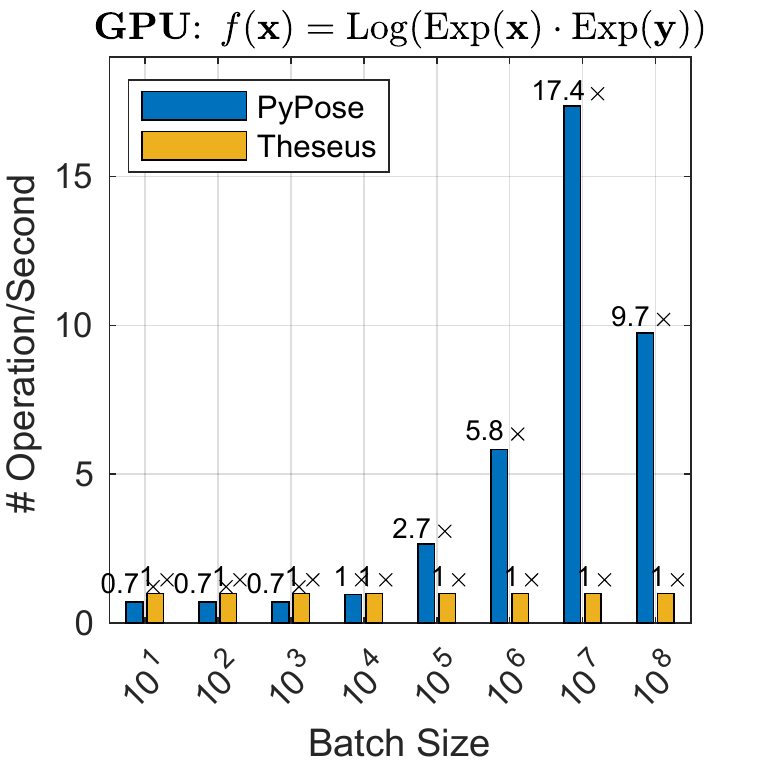}}
	\hfill
	\subfloat[CPU Jacobian runtime.]{\includegraphics[width=0.2\linewidth]{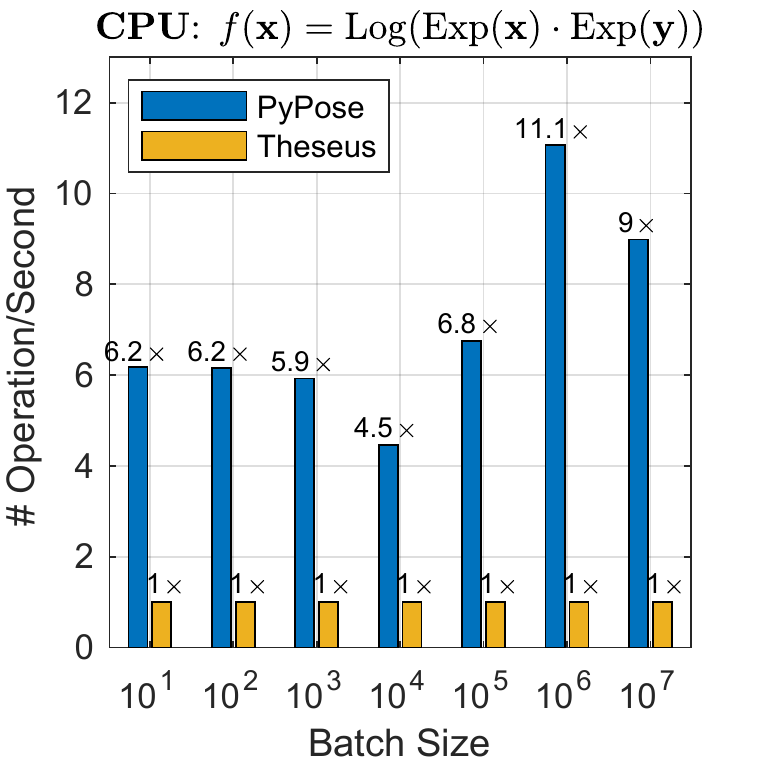}}
	\hfill
	\subfloat[GPU Jacobian runtime.]{\includegraphics[width=0.2\linewidth]{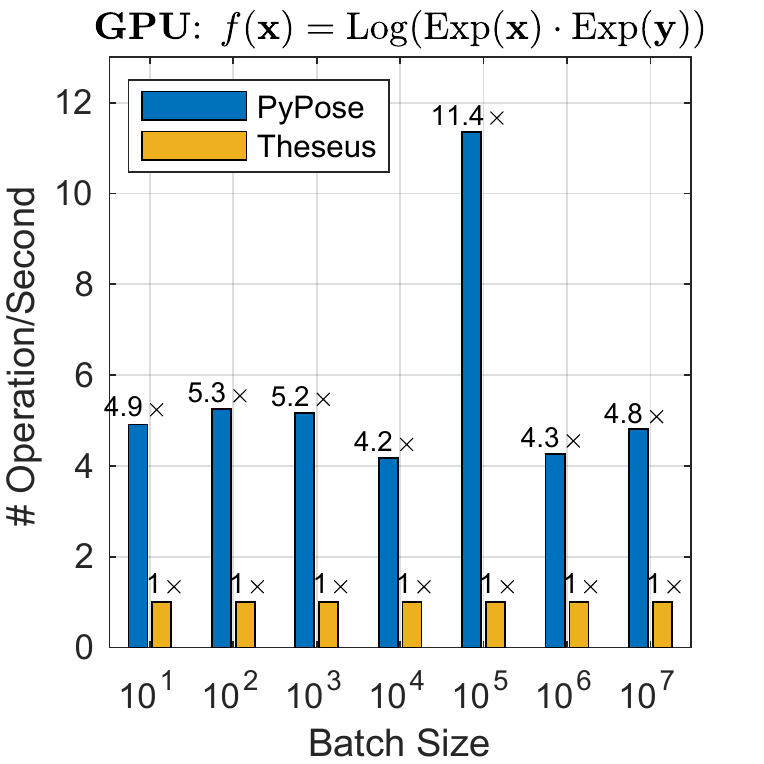}}
	\hfill
	\subfloat[Memory usage.]{\includegraphics[width=0.2\linewidth]{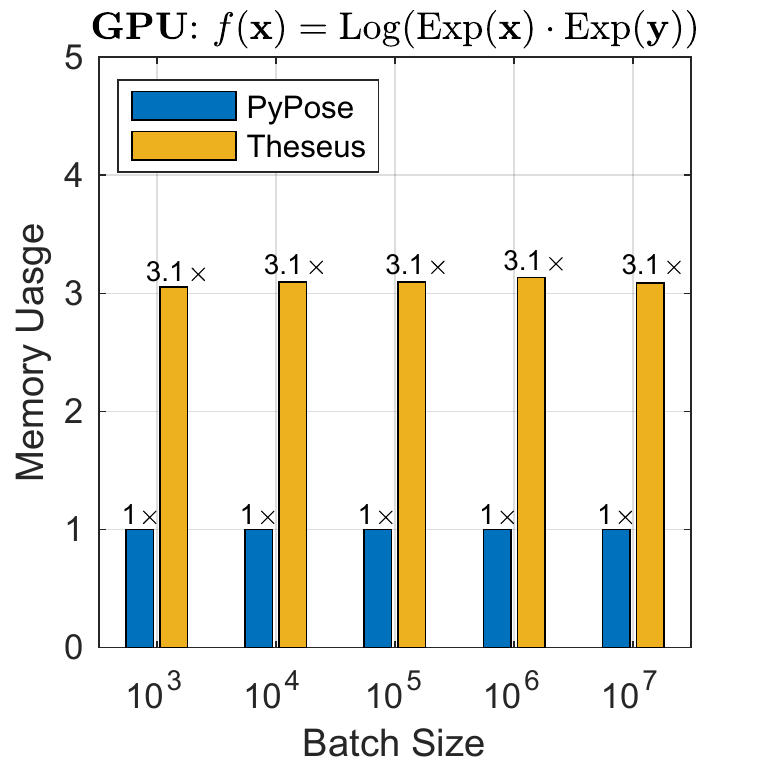}}
	\vspace{5pt}
	\subfloat[CPU runtime.\label{fig:imbalance}]{\includegraphics[width=0.2\linewidth]{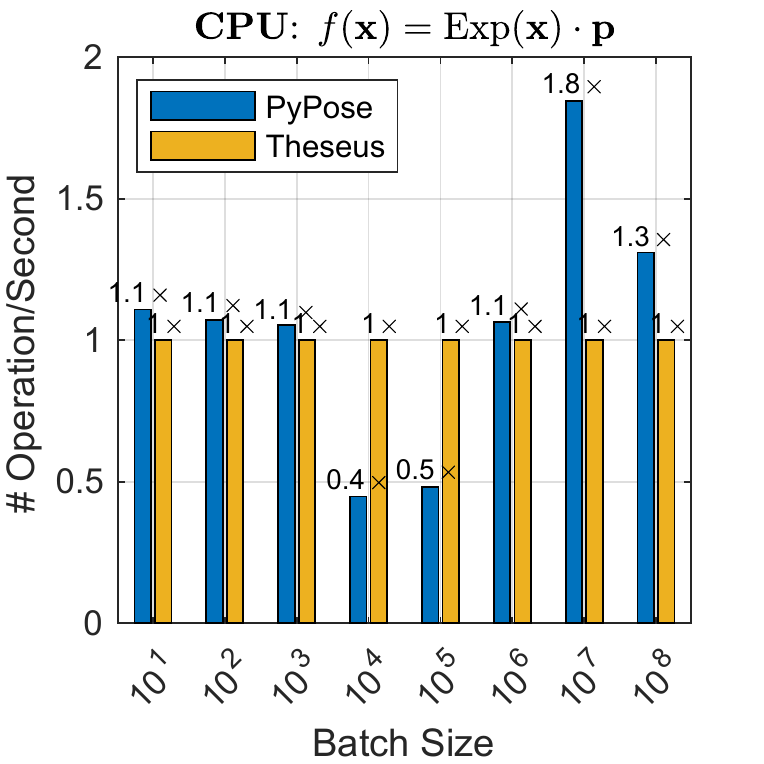}}
	\hfill
	\subfloat[GPU runtime.]{\includegraphics[width=0.2\linewidth]{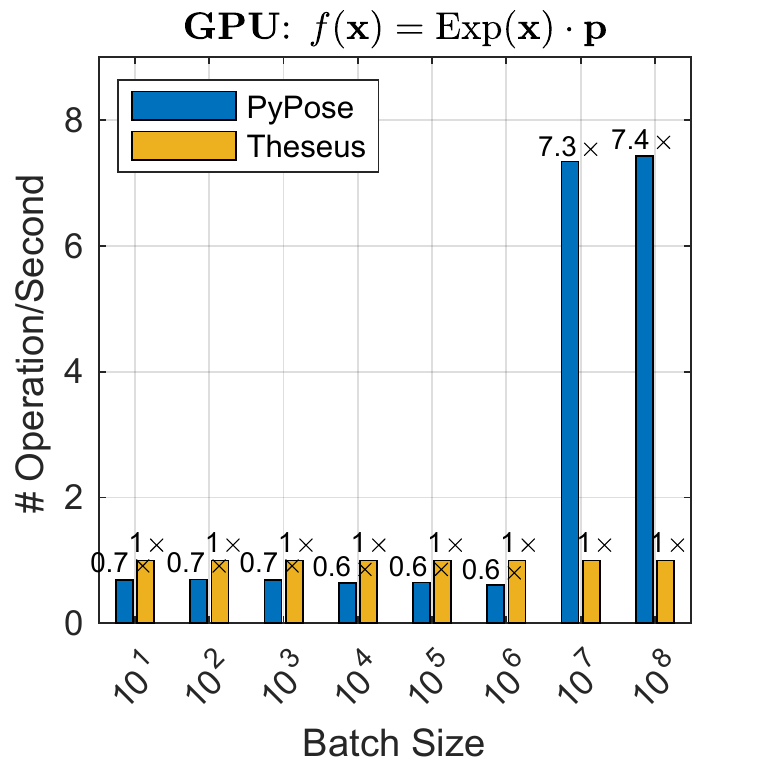}}
	\hfill
	\subfloat[CPU Jacobian runtime.]{\includegraphics[width=0.2\linewidth]{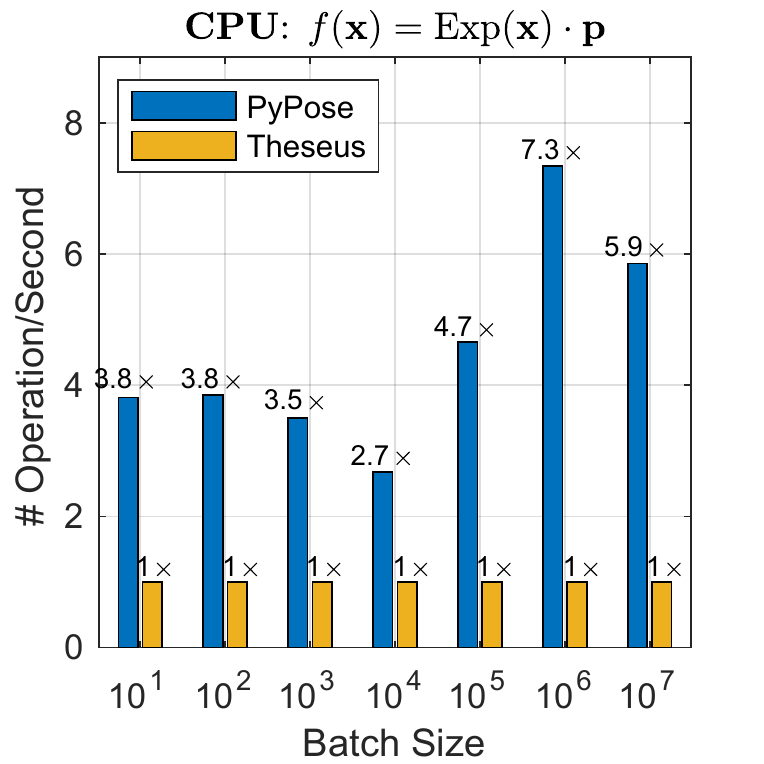}}
	\hfill
	\subfloat[GPU Jacobian runtime.]{\includegraphics[width=0.2\linewidth]{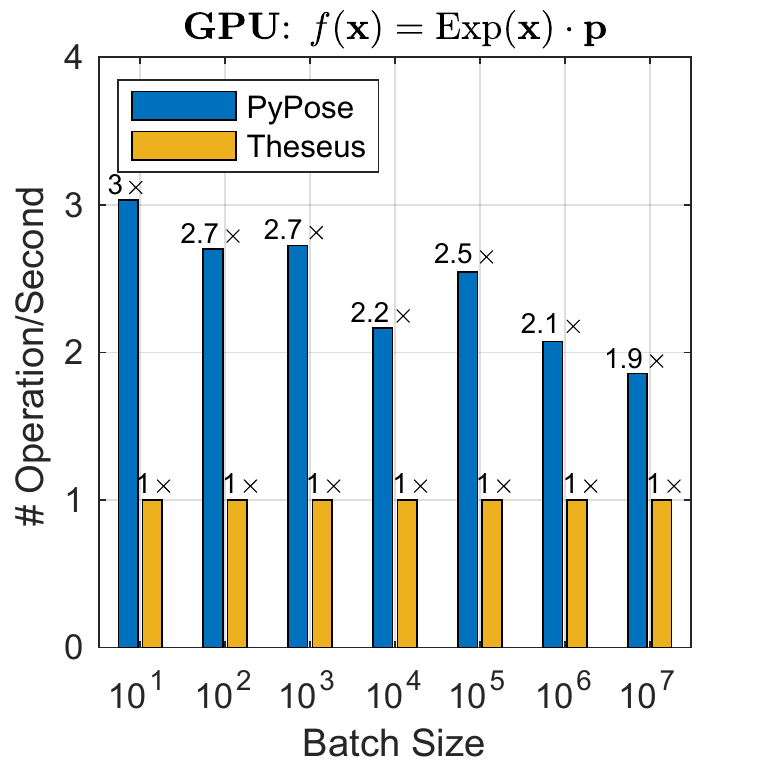}}
	\hfill
	\subfloat[Memory usage.]{\includegraphics[width=0.2\linewidth]{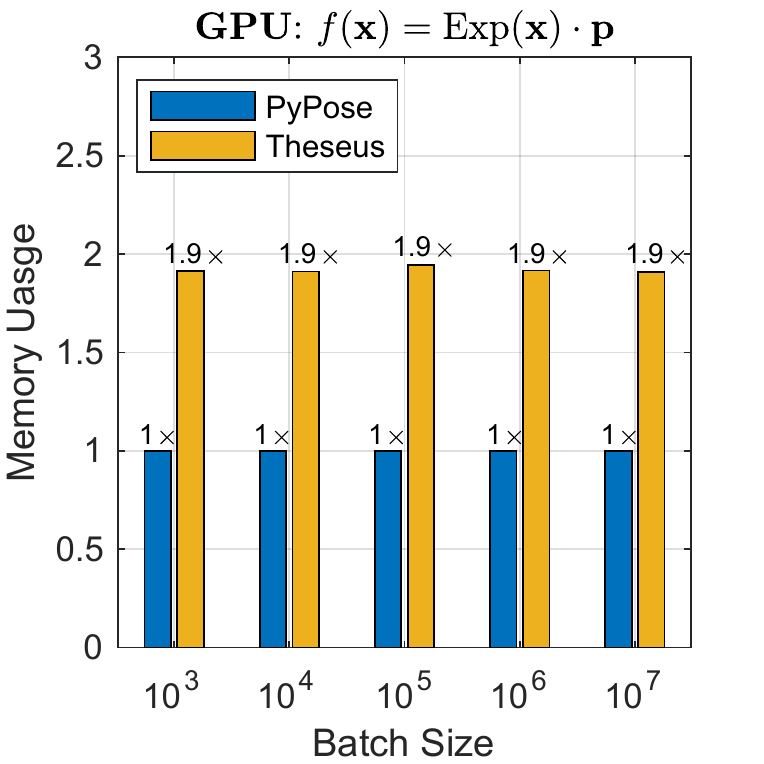}}
	\caption{Efficiency and memory comparison of batched Lie group operations (we take Theseus \cite{pineda2022theseus} performance as $1\times$). Note that Theseus provides multiple ways to calculate Jacobian, and we adopted \texttt{AutoDiffCostFunction} for comparison. This is because this method is used in its optimizers and can reflect the performance of an optimization problem. Theseus also provides analytical Jacobian for some single operators, which could be faster, but one needs to manually compose the Jacobian of each operator for user-defined functions.}
	\label{fig:benchmark}
    \vspace{-5pt}
\end{figure*}

The solution to \eqref{eq:least-square} of an \texttt{LM} algorithm is computed by iteratively updating an estimate $\bm{\theta}_{t}$ via
 $\bm{\theta}_t \leftarrow \bm{\theta}_{t-1} + \bm{\delta}_t$, where the update step $\bm{\delta}_t$ is computed as:
\begin{equation}\label{eq:step}
   \sum_{i}\left(\mathbf{H}_i + \lambda\cdot\mathrm{diag}(\mathbf{H}_i)\right)  \bm{\delta}_t = - \sum_{i}\mathbf{J}_i^T \mathbf{W}_i \mathbf{R}_i,
\end{equation}
where $\mathbf{R}_i = \bm{f}(\bm{\theta}_{t-1}, \bm{x}_i)-\bm{y}_i$ is the $i$-th residual error, $\mathbf{J}_i$ is the Jacobian of $\bm{f}$ computed at $\bm{\theta}_{t-1}$, $\mathbf{H}_i$ is an approximate Hessian matrix computed as $\mathbf{H}_i = \mathbf{J}_i^T\mathbf{W}_i\mathbf{J}_i$, and $\lambda$ is a damping factor.
To find step $\bm{\delta}_t$, one needs a linear \texttt{solver}:
\begin{equation}
    \mathbf{A} \cdot \bm{\delta}_t = \bm{b},
\end{equation}
where $\mathbf{A} = \sum_{i}\left(\mathbf{H}_i + \lambda\cdot\mathrm{diag}(\mathbf{H}_i)\right)$, $\bm{b} = - \sum_{i}\mathbf{J}_i^T \mathbf{W}_i \mathbf{R}_i$.
In practice, the square matrix $\mathbf{A}$ is often positive-definite, so we could leverage standard linear solvers such as Cholesky. If the Jacobian $\mathbf{J}_i$ is large and sparse, we may also use sparse solvers such as sparse Cholesky \cite{chadwick2015efficient} or preconditioned conjugate gradient (PCG) \cite{hestenes1952methods} solver.

In practice, one often introduces robust \texttt{kernel} functions $\rho: \mathbb{R}\mapsto\mathbb{R}$ into \eqref{eq:least-square} to reduce the effect of outliers:
\begin{equation}\label{eq:kernel-least-square}
    \min_{\bm{\theta}} \sum_i 
            \rho\left(\mathbf{R}_i^T \mathbf{W}_i \mathbf{R}_i\right),
\end{equation}
where $\rho$ is designed to down-weigh measurements with large residuals $\mathbf{R}_i$.
In this case, we need to adjust \eqref{eq:step} to account for the presence of the robust kernel. A popular way is to use Triggs' correction \cite{triggs1999bundle}, which is also adopted by the Ceres \cite{AgarwalCeresSolver2022} library.
However, it needs \nth{2}-order derivative of the kernel function $\rho$, which is always negative. This can lead \nth{2}-order optimizers including LM to be unstable \cite{triggs1999bundle}.
Alternatively, PyPose introduces \texttt{FastTriggs} corrector, which is faster yet more stable than Triggs' correction by only involving the \nth{1}-order derivative:
\begin{equation}
    \mathbf{R}_i^\rho = \sqrt{\rho'(c_i)} \mathbf{R}_i,\quad
    \mathbf{J}_i^\rho = \sqrt{\rho'(c_i)} \mathbf{J}_i,
\end{equation}
where $c_i = \mathbf{R}_i^T\mathbf{W}_i\mathbf{R}_i$, $\mathbf{R}_i^\rho$ and $\mathbf{J}_i^\rho$ are the corrected model residual and Jacobian due to the introduction of kernel functions, respectively.
More details about \texttt{FastTriggs} and its proof can be found in the documentation.\footnote{\href{https://pypose.org/docs/main/generated/pypose.optim.corrector.FastTriggs}{https://pypose.org/docs/main/generated/pypose.optim.corrector.FastTriggs}}

A simple \texttt{LM} optimizer may not converge to the global optimum if the initial guess is too far from the optimum. For this reason, we often need other strategies such as adaptive damping, dog leg, and trust region methods \cite{lourakis2005levenberg} to restrict each step, preventing it from stepping ``too far''.
To adopt those strategies, one may simply pass a \texttt{strategy} instance, e.g., \texttt{TrustRegion}, to an optimizer.
In summary, PyPose supports easy extensions for the aforementioned algorithms by simply passing \texttt{optimizer} arguments to their constructor, including \texttt{solver}, \texttt{strategy}, \texttt{kernel}, and \texttt{corrector}.
A list of available algorithms and examples can be found in the documentation.\footnote{\href{https://pypose.org/docs/main/optim}{https://pypose.org/docs/main/optim}}
A sample code showing how to use an optimizer is provided in the supplementary.

\vspace{-2mm}
\section{Experiments}
\label{sec:experiment}
\vspace{-2mm}
In this section, we showcase PyPose's performance and present practical examples of its use for vision and robotics.

\vspace{-1mm}
\subsection{Runtime Efficiency}
\vspace{-2mm}
We report the runtime efficiency of \texttt{LieTensor} and compare it with the state-of-the-art libraries, including LieTorch \cite{teed2021tangent} and Theseus \cite{pineda2022theseus}.
Since PyPose and LieTorch use quaternions and Theseus uses rotation matrices to represent transformation, they are not directly comparable.
To have an extensive and fair comparison, we construct three widely-used functions with the same input/output shapes for all libraries, including
exponential and logarithm mapping $f_1(\mathbf{x}) = \mathrm{Log}(\mathrm{Exp}(\mathbf{x})),~\mathbf{x}\in \mathfrak{so}(3)$, rotation composition $f_2(\mathbf{x}) = \mathrm{Log}(\mathrm{Exp}(\mathbf{x})\cdot\mathrm{Exp}(\mathbf{y})),~\mathbf{x, y}\in\mathfrak{so}(3)$, and rotating points $f_3(\mathbf{x}) = \mathrm{Exp}(\mathbf{x})\cdot \mathbf{p},~\mathbf{x}\in\mathfrak{so}(3), \mathbf{p}\in\mathbb{R}^3$.
We take the number of operations per second for the operation $f_i$ and its Jacobian $J_i = \frac{\partial f_i}{\partial \mathbf{x}}$ as the metric.

\begin{table}[!t]
    \caption{The final error and runtime of optimizers.}
    \vspace{-5pt}
    \label{tab:optimizer}
    \centering
    \resizebox{\linewidth}{!}{
    \begin{tabular}{c|cc|cc|cc|cc}
        \toprule
        \multirow{2}{*}{Batch} & \multicolumn{2}{c|}{$10^1$} & \multicolumn{2}{c|}{$10^2$} & \multicolumn{2}{c|}{$10^3$} & \multicolumn{2}{c}{$10^4$} \\
          & Error & Time  & Error & Time  & Error & Time  & Error & Time \\
         \midrule
        
       \texttt{SGD} & 5.10E+1 & 0.46 & 4.96E+2 & 0.95 & 4.09E+3 & 1.45 & 6.86E+3 & 31.5 \\
        
        \texttt{Adam} & 9.81E+1 & 2.96 & 1.25E+3 & 9.28 & 1.13E+4 & 16.9 & 1.16E+5 & 40.5 \\
        
        \textbf{\texttt{LM}} & \textbf{5.96E-5} & \textbf{0.57} & \textbf{5.80E-4} & \textbf{0.60} & \textbf{3.72E-3} & \textbf{0.92} & \textbf{3.77E-2} & \textbf{51.6} \\

        \bottomrule
    \end{tabular}
    }
\end{table}

\begin{figure}[t]
	\centering
	\subfloat[Before optimization.\label{fig:pgo-before}]{\includegraphics[width=0.5\linewidth]{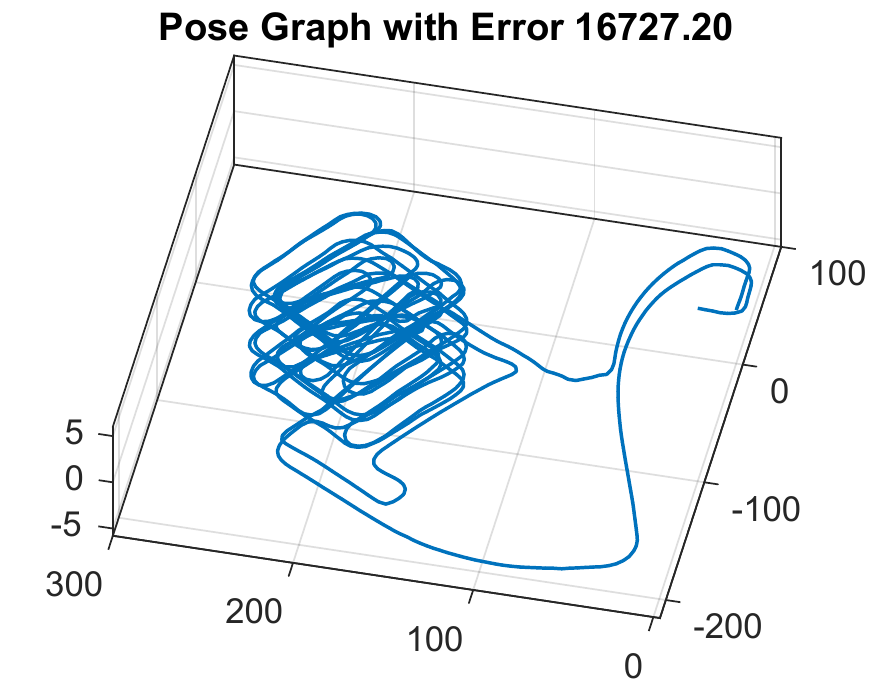}}
	\subfloat[After optimization.\label{fig:pgo-after}]{\includegraphics[width=0.5\linewidth]{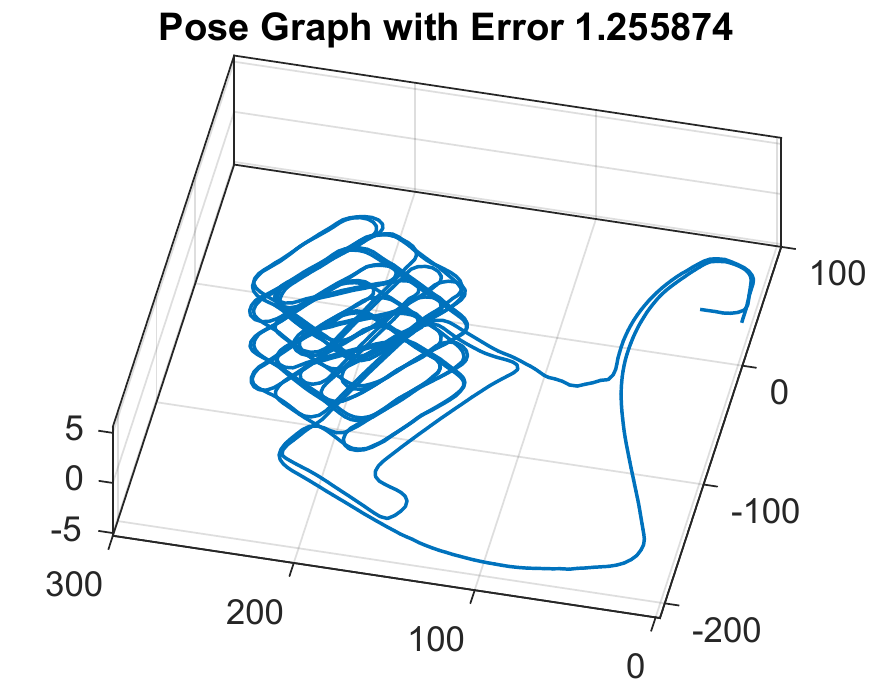}} \\
	\caption{A PGO example ``Garage'' using the \texttt{LM} optimizer. PyPose has the same final error as Ceres \cite{AgarwalCeresSolver2022} and GTSAM \cite{dellaert2012factor}.}
    \vspace{-10pt}
	\label{fig:pgo}
\end{figure}

We first report the performance of calculating dense Jacobian of operator $f_1$ in \fref{fig:jacobian}.
PyPose is up to $37.9\times$ faster on a CPU than Theseus and up to $8.8\times$ faster than LieTorch.
On a GPU, PyPose achieves up to $42.9\times$ faster than Theseus and up to $8.6\times$ faster than LieTorch.

\fref{fig:benchmark} shows the performance of calculating batched operations, where the function inputs $\mathbf{x}$ are assumed to be independent, and the batched Jacobian can be calculated in parallel.
In this case, we have to use ``loop'' for calculating Jacobian with LieTorch; thus, we don't compare with it for fairness, as loop in Python is slow.
On a CPU, PyPose is up to $1.8\times$, $2.8\times$, and $2.2\times$ faster than Theseus for the three operations, respectively.
On a GPU, PyPose shows even better performance, up to $7.4\times$, $32.9\times$, and $17.4\times$ faster.
For computing the batched Jacobian, PyPose achieves up to $7.3\times$, $16.1\times$, and $11.1\times$ on a CPU and $3.0\times$, $14.4\times$, and $11.4\times$ faster on a GPU for the three operations, respectively.

We noticed that PyPose only needs about  $\nicefrac{1}{4}$ to $\nicefrac{1}{2}$ memory space compared to Theseus, which further demonstrates its efficiency.
We report a maximum batch size of $10^7$ because bigger one takes too much memory to handle for our platform.
It is worth mentioning that the runtime usually has a fluctuation within $10\%$, and we report the average performance of multiple tests.
A relative performance drop for certain batch size is seen (\fref{fig:imbalance}), which might be caused by cache miss or imbalanced Linux OS resource managing.

\vspace{-2mm}
\subsection{Optimization}
\vspace{-2mm}
We next report the performance of PyPose's \nth{2}-order optimizer Levenberg-Marquardt (\texttt{LM}) and compare with PyTorch's \nth{1}-order optimizers such as \texttt{SGD} and \texttt{Adam}.
Specifically, we construct a \texttt{Module} to learn transformation inverse, as provided in the PyPose tutorials and report their performance in \tref{tab:optimizer}.
Note that PyPose provides analytical solutions for inversion, and this module is only to benchmark the optimizers.
It can be seen that this problem is quite challenging for \nth{1}-order optimizers, while \nth{2}-order optimizers can find the best solution with a much lower final error quickly, which verifies the effectiveness of PyPose's implementation.

\begin{table}[!t]
    \caption{Mean matching accuracy improves after self-supervised training on TartanAir \cite{wang2020tartanair}.}
    \vspace{-5pt}
    \label{tab:slam-improvement}
    \centering
    \resizebox{\linewidth}{!}{
    \begin{tabular}{ccccc}
      \toprule
      Difficulty & \# Sequence & Original & Trained & Improvement     \\
      \midrule
      Easy & 19 & 21.37\% & 32.04\% & + 49.94\% \\
        Hard & 19 & 18.10\% & 24.50\% & + 35.33\% \\
        Overall & 38 & 19.74\% & 28.27\% & + 43.24\% \\
      \bottomrule
  \end{tabular}
    }
    \vspace{-6pt}
\end{table}

\begin{figure}
    \centering
    \sbox{\measurebox}{%
    \begin{minipage}[b]{.5\linewidth}
        \subfloat[Matching reprojection error.\label{fig:slam-matching}]{\includegraphics[width=\linewidth]{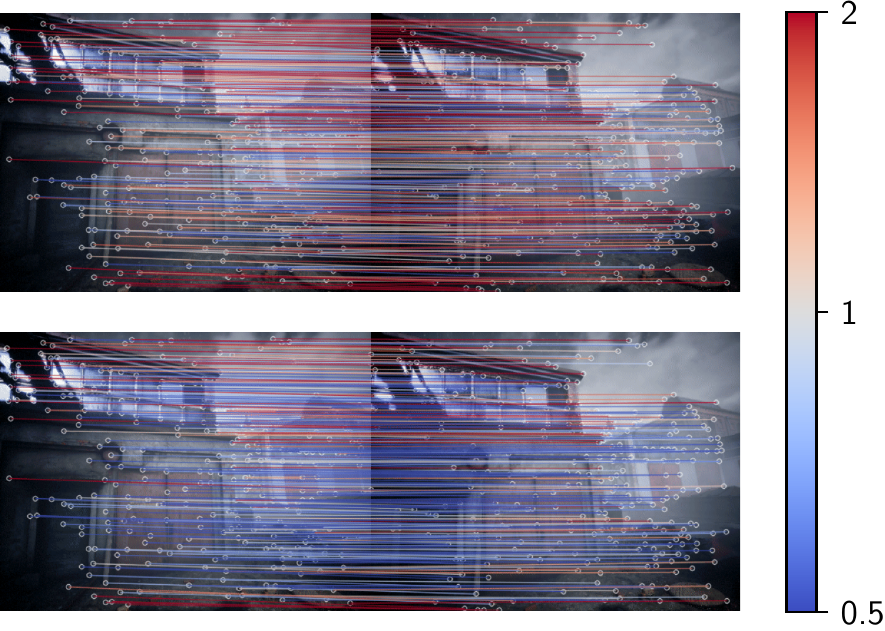}}
    \end{minipage}}
    \usebox{\measurebox}
    \begin{minipage}[b][\ht\measurebox][s]{.47\linewidth}
        \centering
        \subfloat[Trajectory.]{\label{fig:slam-traj}\includegraphics[width=\linewidth]{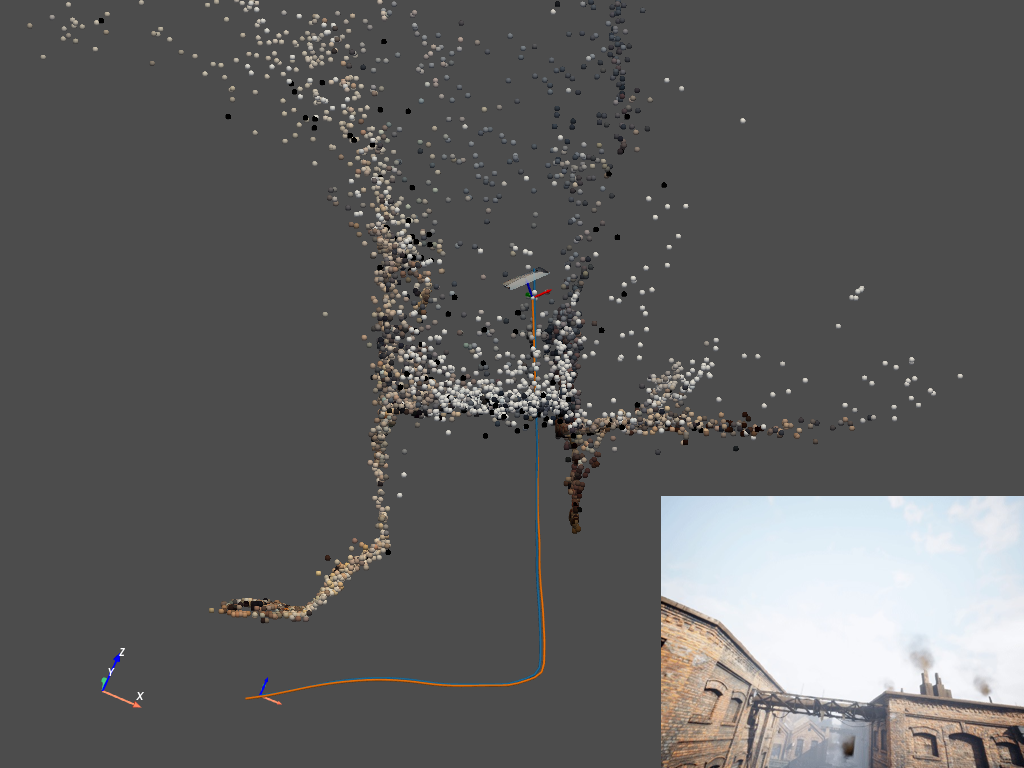}}
    \end{minipage}
    \caption{An example of visual SLAM using PyPose. (a) Matching reprojection error (pixels) (top) is improved after self-supervised tuning using PyPose (bottom). (b) Input sequence (bottom right), resulting trajectory, and point cloud. The estimated and ground truth trajectories are shown in blue and orange, respectively.}
    \label{fig:slam}
    \vspace{-8pt}
\end{figure}

One of the drawbacks of \nth{2}-order optimizers is that they need to compute Jacobians, which requires significant memory and may increase the runtime. Leveraging the sparsity of Jacobian matrices could significantly alleviate this challenge \cite{zach2014robust}.
One example is pose graph optimization (PGO), which is challenging for \nth{1}-order optimizers and also requires pose inversion for every edge in the graph \cite{bai2021sparse}.
We report the performance of PGO using the \texttt{LM} optimizer with a \texttt{TrustRegion} \cite{sun2006optimization} strategy in \fref{fig:pgo}, where the sample is from the g$^2$o dataset \cite{carlone2015lagrangian}.
PyPose achieves the same final error as Ceres \cite{AgarwalCeresSolver2022} and GTSAM \cite{dellaert2012factor}.
An example using PGO for inertial navigation is presented in \sref{sec:imu}.

\vspace{-2mm}
\subsection{Practical Examples} \label{sec:examples}
\vspace{-2mm}

\begin{figure}[t]
	\centering
	\subfloat[Planning runtime.\label{fig:computing-time}]{\includegraphics[height=0.341\linewidth]{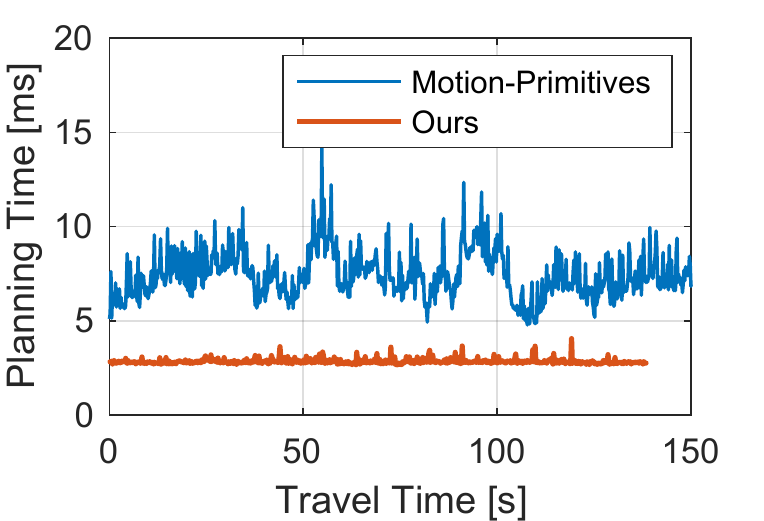}}
	\hfill
	\subfloat[Executed trajectory.\label{fig:trajectory}]{\includegraphics[height=0.325\linewidth]{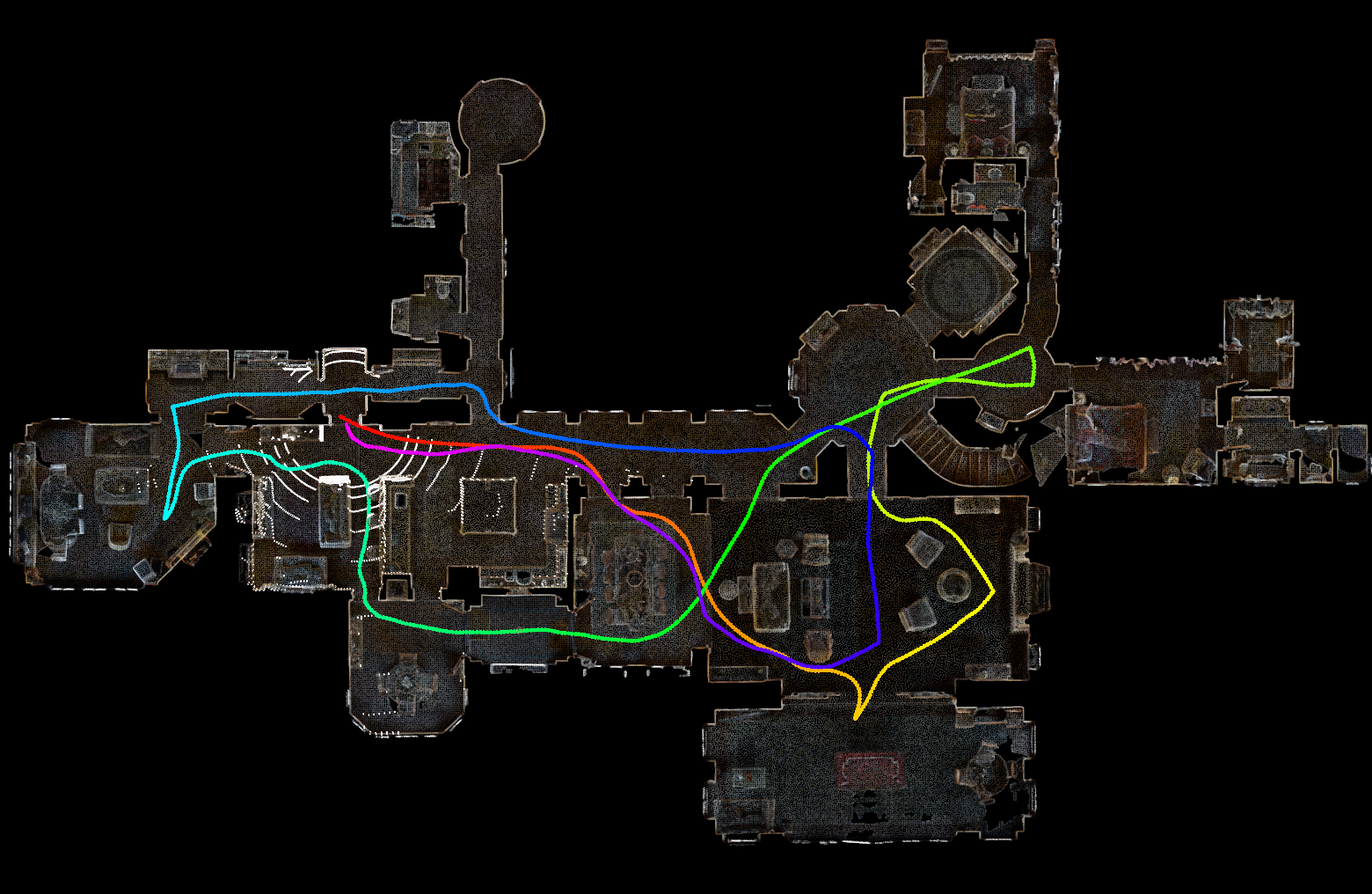}}
	\hfill
    \subfloat[Input depth image.\label{fig:field-test-image}]{\includegraphics[height=0.34\linewidth]{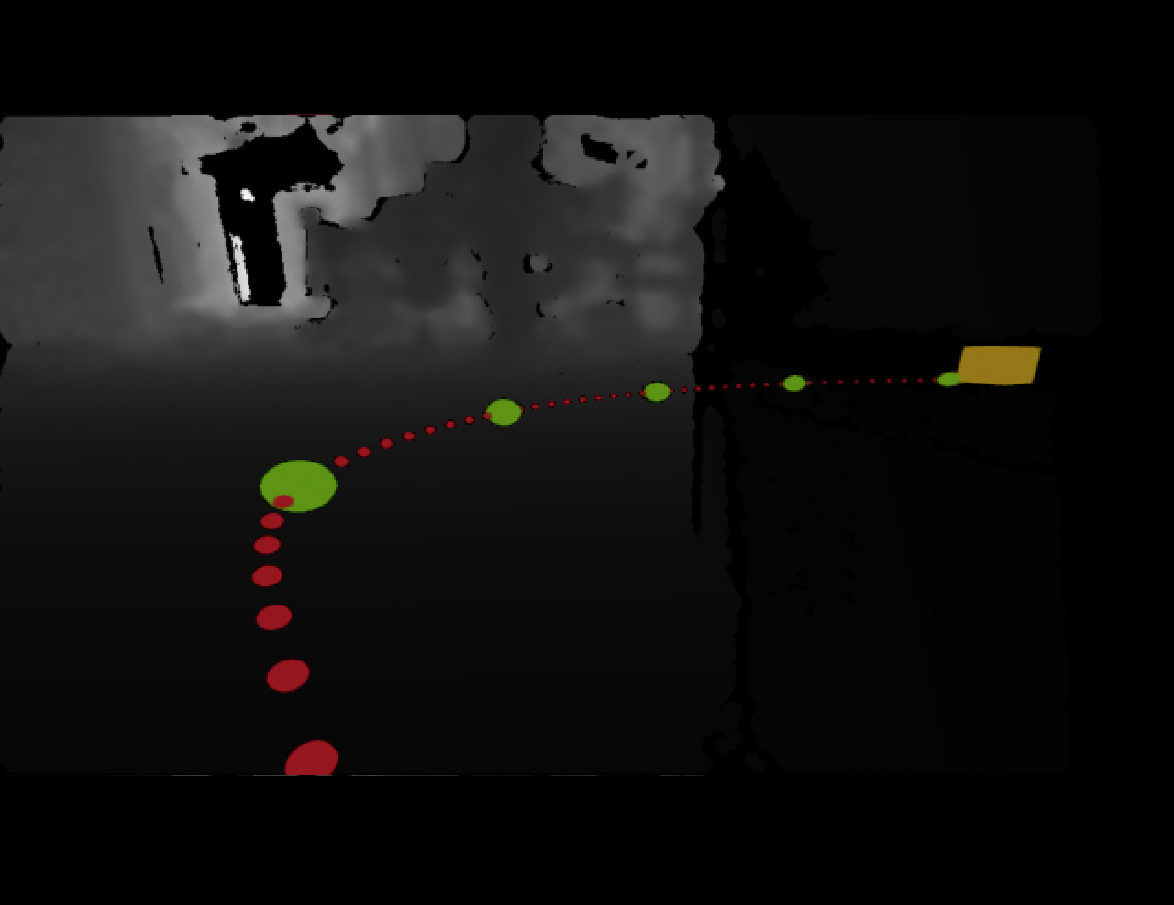}}
    \hfill
    \subfloat[Planning Instance.\label{fig:field-test-traj}]{\includegraphics[height=0.34\linewidth]{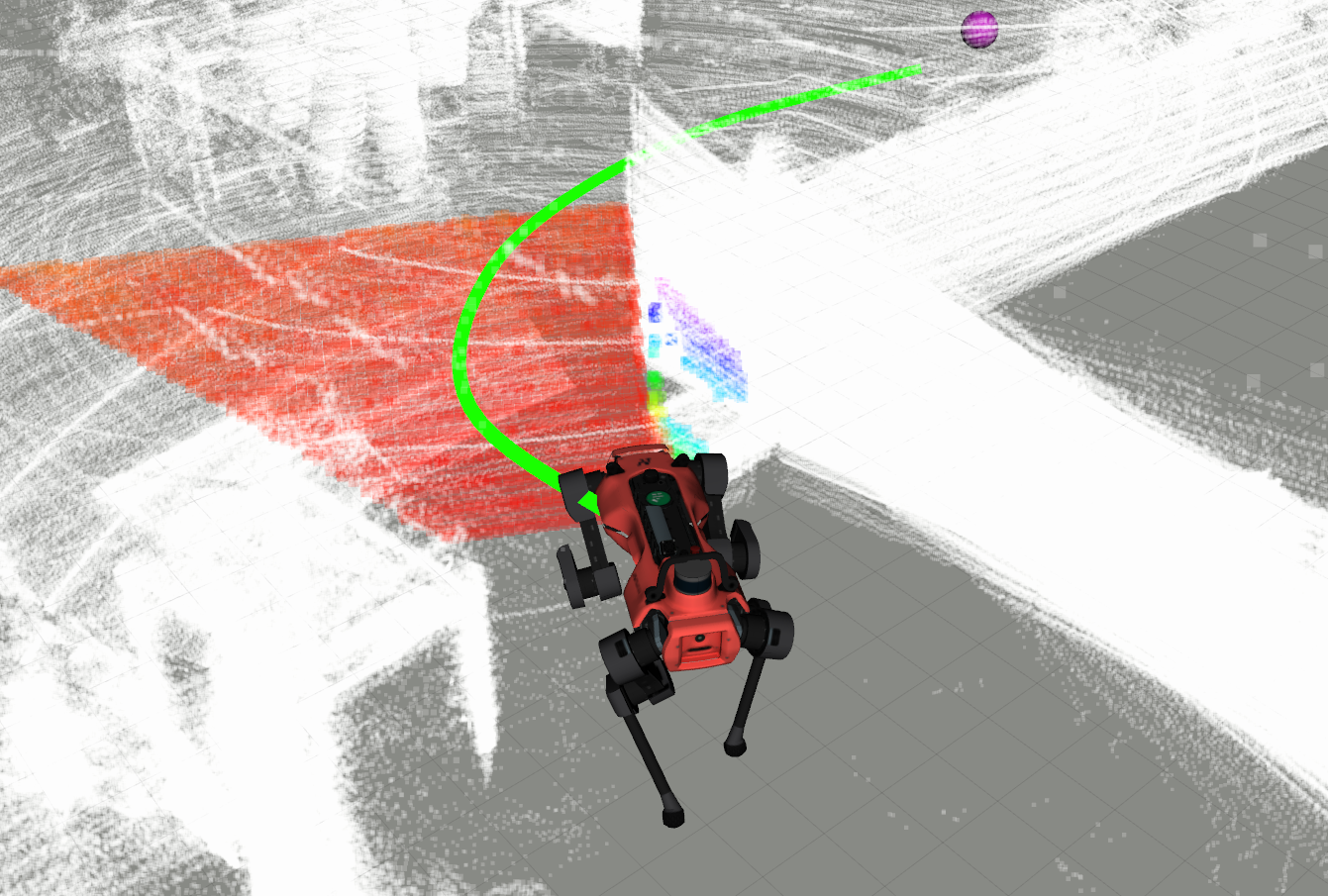}}
	\caption{An example of an end-to-end planning using PyPose library. (a) and (b) show a local planning experiment conducted in the Matterport3D simulated environment. (a) shows the comparison plot of the algorithm planning time. (b) shows the trajectory executed by the proposed method and the outline of the environment. (c) and (d) show an instance of the real-world experiment conducted with ANYmal-legged robot. 
	A kinodynamically feasible trajectory, shown in (d), is directly generated from the input depth image (c). The red dots in (c) show the trajectory projected back into the image frame for visualization.}
	\label{fig:planner}
	\vspace{-10pt}
\end{figure}

\subsubsection{SLAM}
\vspace{-2mm}
To showcase PyPose's ability to bridge learning and optimization tasks, we develop a method for learning the SLAM front-end in a self-supervised manner.
Due to difficulty in collecting accurate camera pose and/or dense depth, few dataset options are available to researchers for training correspondence models offline.
To this end, we leverage the estimated poses and landmark positions as pseudo-labels for supervision.
At the front end, we employ two recent works, SuperPoint \cite{detone2018superpoint} and CAPS \cite{wang2020learning}, for feature extraction and matching, respectively.
Implemented with the \texttt{LM} optimizer of PyPose, the backend performs bundle adjustment and back-propagates the final projection error to update the CAPS feature matcher.
In the experiment, we start with a CAPS network pretrained on MegaDepth \cite{li2018megadepth} and finetune on the TartanAir \cite{wang2020tartanair} dataset with only image input.
We segment each TartanAir sequence into batches of 6 frames and train the CAPS network with the estimated poses.
To assess the effect of unsupervised training, we evaluate the CAPS model on the two-frame matching task and report the results in \tref{tab:slam-improvement}.
It can be seen that the matching accuracy (reprojection error $\leq$ 1 pixel) increased by up to 50\% on unseen sequences after self-supervised training.
We show a visual comparison of the reprojection error distribution before and after training in \fref{fig:slam-matching} and the resulting trajectory and point cloud on a test sequence in \fref{fig:slam-traj}.
While the original model quickly loses track, the trained model runs to completion with an ATE of 0.63 \meter.
This verifies the feasibility of PyPose for optimization in the SLAM backend.

\begin{figure}
	\centering
	\subfloat[Model loss.\label{fig:model_loss}]{\includegraphics[width=0.498\linewidth]{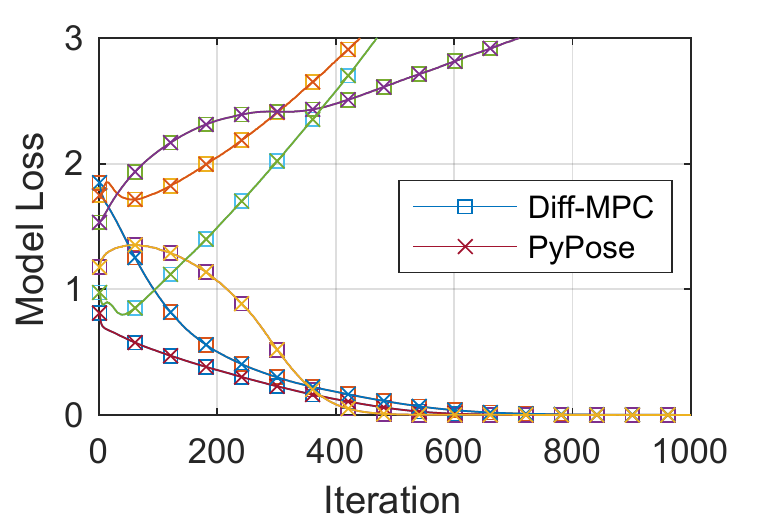}}
	\hfill
	\subfloat[Trajectory
loss.\label{fig:traj_loss}]{\includegraphics[width=0.498\linewidth]{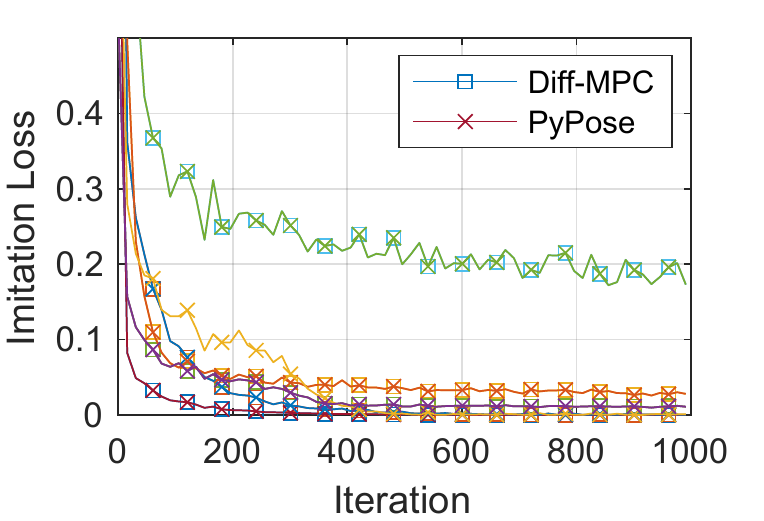}} \\
    \subfloat[Overall runtime.\label{fig:total_time}]{\includegraphics[width=0.498\linewidth]{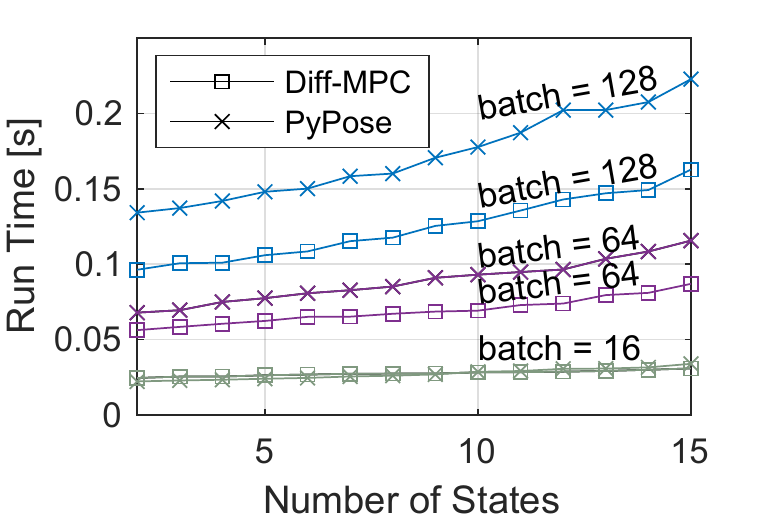}}
    \hfill
    \subfloat[Backwards runtime.\label{fig:backward_time}]{\includegraphics[width=0.498\linewidth]{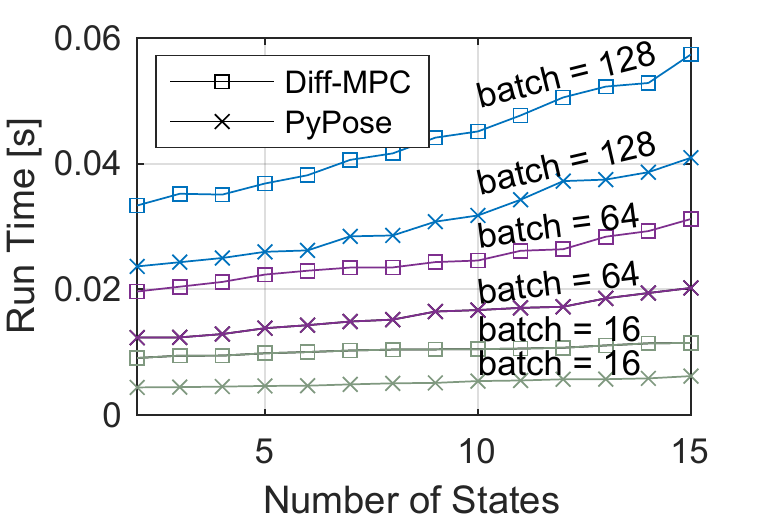}} \\
	\caption{An example of MPC with imitation learning using PyPose library. (a) and (b) show the learning performance comparison of PyPose and Diff-MPC \cite{amos2018differentiable}. The two methods achieve the same learning performance. (c) and (d) show the computation cost comparison, where different colors represent different batch sizes.}
	\vspace{-10pt}
	\label{fig:control}
\end{figure}

\vspace{-3mm}
\subsubsection{Planning}
\vspace{-2mm}
The PyPose library has been used to develop a novel end-to-end planning policy that maps the depth sensor inputs directly into kinodynamically feasible trajectories. Here, we combined learning and trajectory optimization to achieve efficient planning and enforce smoothness and kinodynamic feasibility for the output paths. The \texttt{SE3} LieTensor of PyPose library is utilized to form a differential layer for trajectory optimization. As shown in \fref{fig:computing-time}, our method achieves around $3\times$ speedup on average compared to a traditional planning framework~\cite{cao2022autonomous}, which utilizes a combined pipeline of geometry-based terrain analysis and motion-primitives-based planning~\cite{zhang2020falco}. The experiment is conducted in the Matterport3D~\cite{Matterport3D} environment. The robot follows 20 manually inputted waypoints for global guidance but plans fully autonomously by searching feasible paths and avoiding local obstacles. The trajectory executed by the proposed method is shown in \fref{fig:trajectory}. The efficiency of this method benefits from both the end-to-end planning pipeline and the efficiency of the PyPose library for training and deployment. Furthermore, this end-to-end policy has been integrated and tested on a real-legged robot, ANYmal. A planning instance during the field test is shown in \fref{fig:field-test-traj} using the current depth observation, shown in \fref{fig:field-test-image}.

\vspace{-3mm}
\subsubsection{Control}
\vspace{-2mm}
PyPose integrates the dynamics and control tasks into the end-to-end learning framework. We demonstrate this capability using a learning-based MPC for an imitation learning problem described in Diff-MPC~\cite{amos2018differentiable}, where both the expert and learner employ a linear-quadratic regulator (LQR) that share all information except for linear system parameters and the learner tries to recover the dynamics using only expert controls. 
We treat MPC as a generic policy class with parameterized cost functions and dynamics, which can be learned by automatic differentiating (AD) through LQR optimization.
Unlike the existing method of simply unrolling and differentiating through the optimization iterations~\cite{tamar2017learning}, which incurs significant computational costs for both the forward and backward passes, or the method of computing the derivative analytically by solving the KKT condition of the LQR problem at a fix point~\cite{amos2018differentiable}, which is problem-specific, in PyPose, we use a \textit{problem-agnostic} AD implementation that applies one extra optimization iteration at the optimal point in the forward pass to enable the automatic gradient computation for the backward pass. Thus, our method does not require the computation of gradients for the entire unrolled chain or analytical derivative for the specific problem. 

The preliminary results are shown in \fref{fig:control} for a benchmarking problem~\cite{amos2018differentiable}.
\fref{fig:model_loss} and \ref{fig:traj_loss} show the model and trajectory losses, respectively, of four trajectories with randomly sampled initial conditions. We can see both methods achieve the same learning performance for different sample dynamics. 
We then compared the two methods' overall runtime and backward time. Our backward time is always shorter because the method in~\cite{amos2018differentiable} needs to solve an additional iteration LQR in the backward pass. Overall, when the system is low-dimensional, our method shows a faster learning process. As the system dimension increases, the runtime of our method gets longer.
The overhead is due to the extra evaluation iteration in the forward pass needs to create and maintain variables for the backward pass. However, we noticed that the overall runtime difference is not significant. The advantage of our method is more prominent as more complex optimization problems are involved.

\begin{figure}[t]
	\centering
	\subfloat[Propagated IMU Covariance.]{\includegraphics[width=0.515\linewidth]{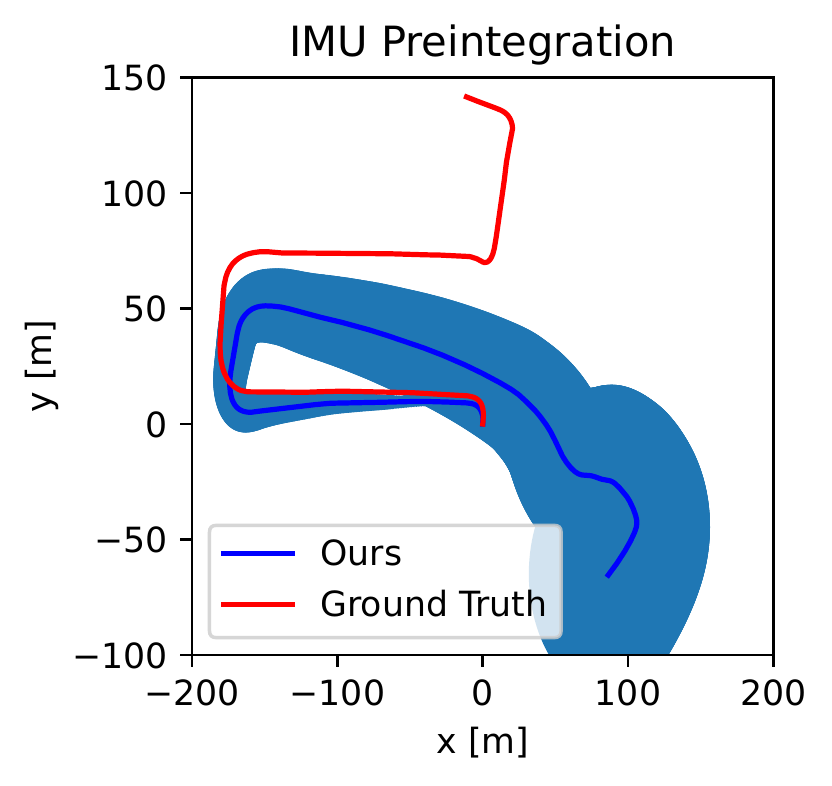}\label{Fig:imuintegra}}
	\subfloat[Odometry After PGO.]{\includegraphics[width=0.485\linewidth]{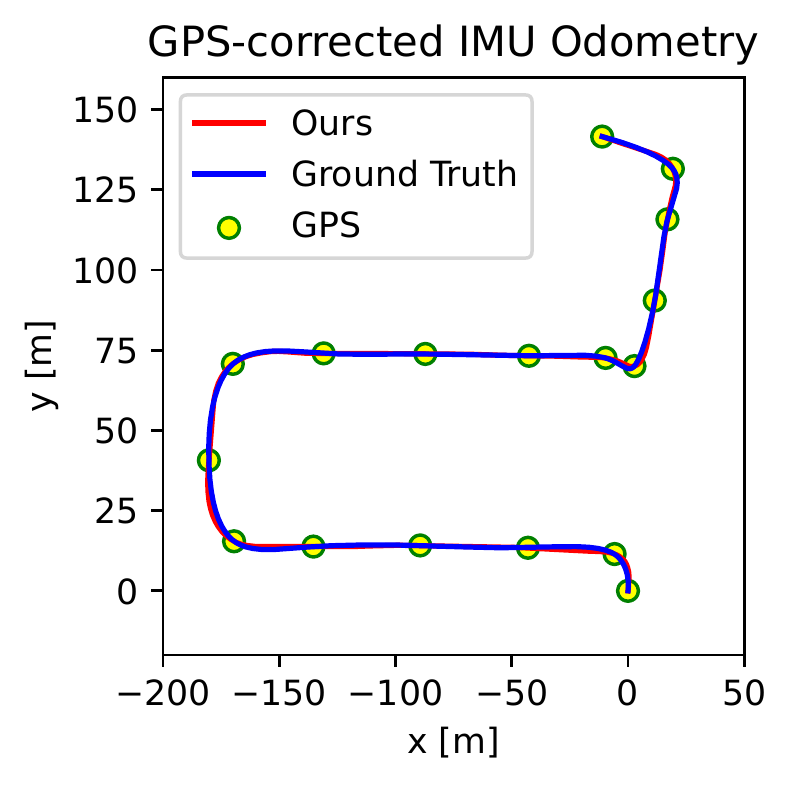}\label{Fig:imupgo}} \\
	\caption{Examples of IMU integration and the IMU-centric PGO on a sequence from the KITTI dataset. 
    Compared with the raw signal, the optimized result is very close to the ground truth.}
    \vspace{-10pt}
\end{figure}

\vspace{-3mm}
\subsubsection{Inertial Navigation} \label{sec:imu}
\vspace{-2mm}
IMU preintegration \cite{forster2015imu} is one of the most popular strategies for many inertial navigation systems, e.g., VIO \cite{qin2018vins}, LIO \cite{zuo2020lic}, and multi-sensor fusion \cite{zhao2021super}.
A growing trend consists in leveraging the IMU's physical prior in learning-based inertial navigation, e.g., IMU odometry network \cite{chen2018ionet}, data-driven calibration \cite{brossard2020denoising, zhang2021imu}, and EKF-based odometry \cite{liu2020tlio, sun2021idol}.
To boost future research, PyPose develops an \texttt{IMUPreintegrator} module for differentiable IMU preintegration with covariance propagation.
It supports batched operation and integration on the manifold space.

\begin{figure}[t]
	\centering
    \includegraphics[width=1\linewidth]{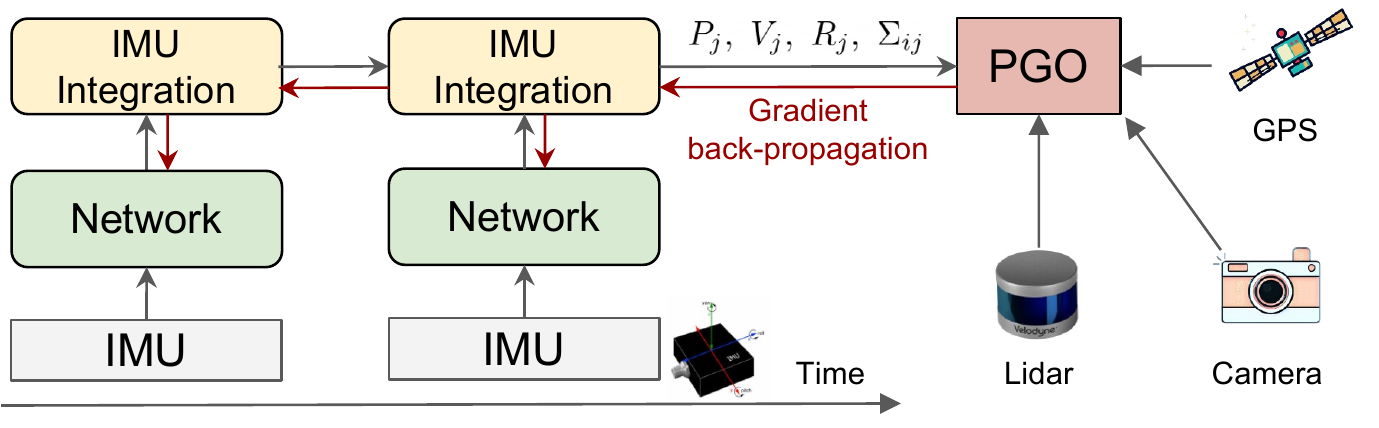}
    \caption{The framework of IMU calibration network using PyPose's \texttt{IMUPreintegrator} with \texttt{LieTensor}.}
	\label{Fig:Integrate}
\end{figure}

\begin{table}[!t]
    \caption{The results of IMU pre-integration in 1\second~(200 frames) on the testing sequences of the EuRoC dataset.}
    \vspace{-5pt}
    \label{tab:imupreinte}
    \centering
    \resizebox{\linewidth}{!}{
    \begin{tabular}{ccccccccccc}
        \toprule
        \multirow{2}{*}{Seq} & \multicolumn{2}{c}{\textbf{MH\_02\_easy}} & \multicolumn{2}{c}{\textbf{MH\_04\_hard}} & \multicolumn{2}{c}{\textbf{V1\_03\_hard}} & \multicolumn{2}{c}{\textbf{V2\_02\_med}} & \multicolumn{2}{c}{\textbf{V1\_01\_easy}}\\
         & pos & rot & pos & rot & pos & rot & pos & rot & pos & rot\\
         \midrule

        Preinte \cite{forster2015imu} & 0.050 & 0.659 & 0.047 & 0.652 & 0.062 & 0.649 & 0.041 & 0.677 & 0.161 & 0.649\\

        Ours & \textbf{0.012} & \textbf{0.012} & \textbf{0.017} & \textbf{0.016} & \textbf{0.025} & \textbf{0.036} & \textbf{0.038} & \textbf{0.054} & \textbf{0.124} & \textbf{0.056}\\
         
        \bottomrule
    \end{tabular}
    }
    \vspace{-10pt}
\end{table}

In \fref{Fig:imuintegra}, we present an IMU integration example on the KITTI dataset \cite{geiger2013vision}, where the blue shadows indicate the propagated IMU covariance.
In this example, the trajectory drifts due to the accumulated sensor noise.
However, as shown \fref{Fig:imupgo}, it can be improved by PGO with GPS (loop closing) for every 5\second~ segment, which also verifies the effectiveness of our \nth{2}-order optimizers.

We next train an IMU calibration network shown in \fref{Fig:Integrate} to denoise the IMU signals, after which we integrate the denoised signal and calculate IMU's state expressed with \texttt{LieTensor} including position, orientation, and velocity.
To learn the parameters in the network, we supervise the integrated pose and back-propagate the gradient through the integrator module.
To validate the effectiveness of this method, we train the network on 5 sequences and evaluate on 5 testing sequences from EuRoC dataset \cite{burri2016euroc} with root-mean-square-error (RMSE) on position (\meter) and rotation (\rad). As shown in \tref{tab:imupreinte}, our method significantly improved orientation and translation compared to the traditional method \cite{forster2015imu}.
This experiment shows that PyPose can benefit future research on the learning-based multi-modal SLAM and inertial navigation systems.

\vspace{-2mm}
\section{Discussion}
\label{sec:conclusion}
\vspace{-2mm}
We present PyPose, a novel pythonic open source library to boost the development of the next generation of robotics. 
PyPose enables end-to-end learning with physics-based optimization and provides an imperative programming style for the convenience of real-world robotics applications.
It is designed to be easy-interpretable, user-friendly, and efficient with a tidy and well-organized architecture.
To the best of our knowledge, PyPose is one of the first comprehensive pythonic libraries covering most sub-fields of robotics where physics-based optimization is involved.

Despite its promising features, PyPose is still in its early stage; its overall stability is not yet mature for practical vision and robotic applications compared to PyTorch.
Moreover, its efficiency is still not comparable with C(++)-based libraries, although it provides an alternative solution for robot learning-based applications.
In the future, we will keep adding more features, including sparse block tensors, sparse Jacobian computation, and constrained optimization.

{\small
\bibliographystyle{ieee_fullname}
\bibliography{sections/06_references, sections/chen}
}

\ifreview 
\clearpage \appendix
\label{sec:appendix}

\section{Sample Code of \textnormal{\texttt{LieTensor}}} \label{app:lietensor}
The following code sample shows how to rotate random points and compute the gradient of batched rotation.

\begin{python}
>>> import torch, pypose as pp

>>> # A random so(3) LieTensor
>>> r = pp.randn_so3(2, requires_grad=True)
so3Type LieTensor:
tensor([[ 0.1606,  0.0232, -1.5516],
        [-0.0807, -0.7184, -0.1102]],
        requires_grad=True)

>>> R = r.Exp() # Equivalent to: R = pp.Exp(r)
SO3Type LieTensor:
tensor([[ 0.0724,  0.0104, -0.6995,  0.7109],
        [-0.0395, -0.3513, -0.0539,  0.9339]],
        grad_fn=<AliasBackward0>)

>>> p = R @ torch.randn(3) # Rotate random point
tensor([[ 0.8045, -0.8555,  0.5260],
        [ 0.3502,  0.8337,  0.9154]],
        grad_fn=<ViewBackward0>)

>>> p.sum().backward()     # Compute gradient
>>> r.grad                 # Print gradient
tensor([[-0.7920, -0.9510,  1.7110],
        [-0.2659,  0.5709, -0.3855]])
\end{python}

\newpage

\section{Sample Code of \textnormal{\texttt{Optimizer}}} \label{app:optimization}

We show how to estimate batched transform inverse by a \nth{2}-order optimizer. Two usage options for a \texttt{scheduler} are provided, each of which can work independently.

\begin{python}
>>> from torch import nn
>>> import torch, pypose as pp
>>> from pypose.optim import LM
>>> from pypose.optim.strategy import Constant
>>> from pypose.optim.scheduler \
    import StopOnPlateau

>>> class InvNet(nn.Module):

        def __init__(self, *dim):
            super().__init__()
            init = pp.randn_SE3(*dim)
            self.pose = pp.Parameter(init)

        def forward(self, input):
            error = (self.pose @ input).Log()
            return error.tensor()

>>> device = torch.device("cuda")
>>> input = pp.randn_SE3(2, 2, device=device)
>>> invnet = InvNet(2, 2).to(device)
>>> strategy = Constant(damping=1e-4)
>>> optimizer = LM(invnet, strategy=strategy)
>>> scheduler = StopOnPlateau(optimizer,
                              steps=10,
                              patience=3,
                              decreasing=1e-3,
                              verbose=True)

>>> # 1st option, full optimization
>>> scheduler.optimize(input=input)

>>> # 2nd option, step optimization
>>> while scheduler.continual():
        loss = optimizer.step(input)
        scheduler.step(loss)
\end{python}

\fi

\ifarxiv 
\clearpage
 
\fi


\end{document}